\documentclass{article}

\usepackage[preprint]{neurips_2025}

\usepackage{mathtools,leftindex,tensor}
\usepackage[version=4]{mhchem}
\usepackage{footmisc}
\usepackage[T1]{fontenc}    
\usepackage{hyperref}       
\usepackage{url}            
\usepackage{booktabs}       
\usepackage{amsfonts, amsmath}       
\usepackage{amsthm}
\usepackage{nicefrac}       
\usepackage{microtype}      
\usepackage{lipsum}		
\usepackage{graphicx}
\usepackage{subcaption} 
\usepackage{natbib}
\usepackage{doi}
\usepackage{xcolor}
\usepackage{caption}

\theoremstyle{plain} 

\newtheorem{definition}{Definition}
\newtheorem{hypothesis}{Hypothesis}
\newtheorem{theorem}{Theorem}
\newtheorem*{theorem_1}{Theorem 1}
\newtheorem*{theorem_2}{Theorem 2}

\usepackage[utf8]{inputenc} 
\usepackage[T1]{fontenc}    
\usepackage{hyperref}       
\usepackage{url}            
\usepackage{booktabs}       
\usepackage{amsfonts}       
\usepackage{nicefrac}       
\usepackage{microtype}      
\usepackage{xcolor}         
\usepackage{enumitem}

\title{A Theory of Machine Understanding via the Minimum Description Length Principle}

\author{%
  Canlin Zhang \\
  Independent \\
  \texttt{canlingrad@gmail.com} \\
  \And
  Xiuwen Liu \\
  Department of Computer Science \\
  Florida State University\\
  Tallahassee, FL 32306 \\
  \texttt{liux@cs.fsu.edu} \\
}

\begin{document}

\maketitle


\begin{abstract}
Deep neural networks trained through end-to-end learning have achieved remarkable success across various domains in the past decade. However, the end-to-end learning strategy, originally designed to minimize predictive loss in a black-box manner, faces two fundamental limitations: the struggle to form explainable representations in a self-supervised manner, and the inability to compress information rigorously following the Minimum Description Length (MDL) principle. These two limitations point to a deeper issue: an end-to-end learning model is not able to \textit{understand} what it learns. In this paper, we establish a novel theory connecting these two limitations. We design the Spectrum VAE, a novel deep learning architecture whose minimum description length (MDL) can be rigorously evaluated. Then, we introduce the concept of latent dimension combinations, or what we term spiking patterns, and demonstrate that the observed spiking patterns should be as few as possible based on the training data in order for the Spectrum VAE to achieve the MDL. Finally, our theory demonstrates that when the MDL is achieved with respect to the given data distribution, the Spectrum VAE will naturally produce explainable latent representations of the data. In other words, explainable representations—or \textit{understanding}—can emerge in a self-supervised manner simply by making the deep network obey the MDL principle. In our opinion, this also implies a deeper insight: \textit{To understand is to compress}. At its core, our theory advocates for a shift in the training objective of deep networks: not only to minimize predictive loss, but also to minimize the description length regarding the given data. That is, a deep network should not only learn, but also understand what it learns. This work is entirely theoretical and aims to inspire future research toward self-supervised, explainable AI grounded in the MDL principle.
\end{abstract}

\section{Introduction}

Deep neural networks have revolutionized machine learning through end-to-end learning \citep{lecun2015deep, wang2020recent}. CNNs, RNNs, and LSTMs have advanced computer vision, language processing, and time-series analysis \citep{lecun1998gradient_initial_cnn, hochreiter1997long, goodfellow2014generative_adversarial_network_GAN, he2016deep}, enabling breakthroughs in image classification, speech recognition, and sequence modeling \citep{deng2009imagenet, graves2006connectionist, sutskever2014sequence}. The Transformer architecture \citep{AttentionIsAll} created a paradigm shift, becoming the foundation for large language models like GPT \citep{brown2020language} and BERT \citep{Devlin_BERT}. Similarly, U-Net \citep{ronneberger2015u} has been crucial for generative AI \citep{feuerriegel2024generative}, particularly in diffusion models for image and video generation \citep{ho2020denoising, song2020score}. These advances demonstrate how end-to-end learning with massive datasets creates versatile deep networks.


Despite these successes, end-to-end learning—originally designed to minimize predictive loss in a black-box manner—exhibits two significant deficiencies. First, networks trained in this manner struggle to form explicit and explainable representations of the data in a self-supervised way \citep{angelov2021explainable}, whereas such a self-supervised learning process is crucial for understanding the physical world \citep{wright2022deep}. Second, according to the Minimum Description Length (MDL) principle, the optimal model should provide the shortest description on the data \citep{hansen2001model}. Yet no end-to-end learning model, including autoencoders \citep{berahmand2024autoencoders}, has been rigorously designed to minimize the description length with respect to the given data distribution.

In this paper, we propose a theory discovering that these two issues are fundamentally connected: By minimizing the description length of a deep learning model with respect to the given data distribution, the model will naturally form explainable latent representations of the data. In other words, explainable representations can be achieved in a self-supervised manner, simply by making the deep neural network obey the MDL principle. Here are our detailed contributions:

\begin{itemize}[leftmargin=12pt]
    \item We establish a theoretical framework for rigorously evaluating the minimum description length (MDL) of a specific model we designed, called the \textit{Spectrum VAE}. To the best of our knowledge, this is the first deep learning architecture designed to minimize the description length.
    \item We introduce the concept of latent dimension combination, or what we term \textit{spiking pattern}, and provide the first analysis of their role in achieving MDL. In a Spectrum VAE, latent dimensions are either zero or positive (\textit{spiking}), forming a \textit{spectrum}. The combination of spiking latent dimensions in a spectrum is referred to as a \textit{spiking pattern}. We demonstrate that, to minimize the description length of a Spectrum VAE on the given data, the observed spiking patterns should be as few as possible. In other words, we seek for \textit{combination sparsity} in latent representations. 
    \item We show that when the Spectrum VAE achieves the MDL, each spiking pattern has to appropriately represent one class of data. That is, the Spectrum VAE will produce explainable latent representations of the data, or \textit{understand} the underlying data structure, in a self-supervised manner, simply by obeying the MDL principle. In our opinion, therefore, \textit{understanding means to represent the acquired information by as small an amount of information as possible}. 
\end{itemize}

More briefly, we believe that \textit{to understand is to compress}. A deep neural network should be trained to not only minimize its predictive loss, but also minimize the description length on the given data. That is, a deep learning model should not only learn, but also understand what it learns.

This paper does not include experimental results. Instead, we focus solely on presenting our theory. In the following part, we will first introduce related work in Section \ref{related_work}. Then, section \ref{main_theory} presents our main theory, including the designed architecture of a Spectrum VAE, the concept of spiking patterns, the way to calculate the description length of a Spectrum VAE on the given data, and the evaluation of a Spectrum VAE when it achieves the MDL. After that, we conclude this paper in Section \ref{conclusion}.

\section{Related Work}\label{related_work}


Variational Autoencoders (VAEs) encode input data $\mathbf{x} \in \mathbb{R}^D$ into a latent Gaussian distribution parameterized by a mean and variance vector \citep{kingma2013_vae_paper}. A latent sample $\mathbf{z}$ is then drawn using the reparameterization trick \citep{kingma2015_VAE_reparameter_trick} based on the generated mean and variance vectors. However, in this formulation, $\mathbf{z}$ is sampled from an input-dependent Gaussian distribution, whose range and scale vary across data points. As a result, it is difficult to apply fixed grid quantization to $\mathbf{z}$, which is however an essential step for rigorously evaluating the Minimum Description Length (MDL) of the model \citep{blier2018description_DL}.

To encourage more structured latent spaces, the $\beta$-VAE \citep{higgins2017beta} introduces a hyperparameter $\beta$ to scale the KL divergence term, promoting disentanglement \citep{carbonneau2022measuring, chen2018isolating}. While this improves interpretability, it assumes a small number of independent semantic factors. In real-world data, however, the number of concepts is often too large to assign one dimension per factor, as seen in large-scale systems like CLIP \citep{radford2021learning}.

Discrete VAEs such as VQ-VAE \citep{van2017neural} address some of these issues by quantizing the latent space using a learned codebook, offering benefits like improved compression and avoidance of posterior collapse \citep{razavi2019preventing_2}. Extensions such as VQ-VAE-2 \citep{razavi2019generating} and dVAE \citep{rolfe2016discrete} enhance these models with hierarchical or improved training strategies. Yet, they still lack a framework for rigorously calculating MDL.

In contrast, our work proposes the \textit{Spectrum VAE}, a new architecture whose MDL can be rigorously calculated. Instead of assigning one latent dimension per concept, we apply combinations of spiking latent dimensions—referred to as \textit{spiking patterns}. This approach potentially enables a small latent space to represent a large number of categories by encoding each category through a distinct spiking pattern. We will provide a detailed discussion in the following section.

\section{Main Theory}\label{main_theory}

In this section, we first introduce the architecture of a Spectrum VAE. Then, to measure the number of bits required to encode a spiking pattern (the combination of spiking latent dimensions), we introduce the concept of $U$-robustness. After that, we propose two hypotheses based on the generalization ability of deep networks. To be specific, we claim that if the observed spiking patterns are few enough, or `dominant enough', based on the training data, we are then unlikely to observe new spiking patterns on the test data. Accordingly, the way to evaluate the minimum description length (MDL) of a Spectrum VAE under given conditions is described. Finally, we demonstrate that a Spectrum VAE has to form explainable latent representations of the data when the MDL is achieved.

\subsection{Spectrum Variational Autoencoder}\label{3.1}

As we mentioned, we aims at accurately calculating the description length of the latent representations in our Variational Autoencoder (VAE) \citep{kingma2013_vae_paper}. Given the input sample $\mathbf{x}\in\mathbb{R}^D$, suppose the encoder (parameterized by $\phi$) in our VAE first produces a preliminary latent vector $\mathbf{z}_{\text{pre}}\in\mathbb{R}^K$. Then, based on two given parameters $0 < a < b$, we truncate each $z_{pre,k}$, the $k$-th dimension of $\mathbf{z}_{\text{pre}}$, for $k=1,\dots,K$: A value below $a$ is set to zero, and a value above $b$ is capped at $b$. This results in a vector $\mathbf{z}\in\mathbb{R}^K$, where the $k$-th dimension $z_k$ is given by:
\begin{align}\label{format_1}
z_k= \begin{cases} 
z_{pre, k} & \text{if } a \leq z_{pre, k} \leq b, \\
b & \text{if } z_{pre, k} > b, \\
0 & \text{if } z_{pre, k} < a.
\end{cases} \ \ \ \text{for } \ k=1,\dots, K.
\end{align}
One can see that there is a discontinuity at value $a$ when mapping $z_{pre, k}$ to $z_k$: When $z_{pre, k}$ reaches $a$ from below, $z_k$ will skip from 0 to $a$ at once. When $z_{pre, k}$ falls below $a$ from above, $z_k$ will skip from $a$ to 0 in a sudden. We design this discontinuity inspired by spiking neural networks (SNN) \citep{sengupta2019going}. Reasons behind this design are discussed in Appendix \ref{appen_a}.

The obtained latent vector $\mathbf{z}\in\mathbb{R}^K$ will be the final output of the encoder, denoted as $\mathbf{z}=\phi(\mathbf{x})$. That is, $\mathbf{z}$ plays the role of $\boldsymbol{\mu}_{\phi}(\mathbf{x})$ in a typical VAE. We can see that $a\leq z_k \leq b$ for each latent dimension $k=1,\dots, K$. The decoder parameterized by $\theta$ will then produce a reconstructed sample $\mathbf{\tilde{x}}=\theta(\mathbf{z})$. Following the routine, we use mean square error (MSE) \citep{hodson2021_MSE} (i.e., the $L_2$ distance between two vectors \citep{nachbar2017vector_sp_norm}), denoted as $\|\mathbf{x}-\mathbf{\tilde{x}}\|_2$, to measure the reconstruction error. A lower MSE means a higher reconstruction fidelity.

Given the latent vector $\mathbf{z}\in\mathbb{R}^K$, we say that $\mathbf{z}$ \textbf{spikes} on the latent dimension $k$ if $z_k\geq a$ (or equivalently, if $z_k > 0$). Accordingly, we call $a$ as the \textbf{spiking threshold} and $b$ as the \textbf{spiking bound}. Since all the spiking and non-spiking latent dimensions in $\mathbf{z}$ arrange like a spectrum, we call $\mathbf{z}\in\mathbb{R}^K$ a \textbf{spectrum}. Multiple spectrum are called \textbf{spectra}. Finally, we call our model a \textbf{Spectrum Variational Autoencoder}, simplified as Spectrum VAE.

In a spectrum $\mathbf{z}\in\mathbb{R}^K$, when a latent dimension equals zero, it cannot carry information \citep{markon2006information}. Thus, information is conveyed only through: (1) the specific pattern, or combination, of spiking latent dimensions, and (2) the precise values of spiking latent dimensions. This spectrum-based latent representation enables us to clearly describe the minimum description length (MDL) \citep{grunwald2007minimum_MDL_book} of our model, as shown in the following parts.


\subsection{$U$-robustness}\label{3.2}
Suppose we have a Spectrum VAE with encoder and decoder parameterized by $\phi$ and $\theta$, respectively. Given a data sample $\mathbf{x}\in\mathbb{R}^D$, suppose the encoder produces a spectrum $\mathbf{z}=\phi(\mathbf{x})\in\mathbb{R}^K$ using spiking threshold $a$ and spiking bound $b$, so that $a\leq z_k \leq b$ for $k=1,\dots, K$. Then, suppose $\mathbf{\tilde{x}}=\theta(\mathbf{z})$ is the reconstructed sample by the decoder.

From all the $K$ latent dimensions, suppose we select and fix $L$ specific ones $k_1, k_2,\dots, k_L$. We call this latent dimension combination $\{k_1,\dots, k_L\}$ a \textbf{pattern}, denoted as $\mathcal{P}=\{k_1,\dots, k_L\}$. 

Now, we ignore the encoder $\phi$ in our Spectrum VAE. For each latent dimension $k_l\in\mathcal{P}$, we give ourself the freedom to choose any possible value $a\leq z_{k_l} \leq b$, without considering whether it can be achieved by the encoder and input samples. 
Also, for each latent dimension $k_l\in\mathcal{P}$, suppose we have a uniform distribution $\mathcal{U}(-\alpha_{k_l}, \alpha_{k_l})$ with a boundary $\alpha_{k_l}>0$ \citep{casella2024statistical}. We then add to $z_{k_l}$ a random scalar $\epsilon_{k_l}$ generated from $\mathcal{U}(-\alpha_{k_l}, \alpha_{k_l})$. But if $z_{k_l}+\epsilon_{k_l}$ goes below $a$ (when $z_{k_l}$ is close to $a$) or exceeds $b$ (when $z_{k_l}$ is close to $b$), we truncate its value back to $[a,b]$ again. That is:
\begin{align}\label{format_2}
\tilde{z}_{k_l}= \begin{cases} 
z_{k_l}+\epsilon_{k_l} & \text{if } a \leq z_{k_l}+\epsilon_{k_l} \leq b, \\
b & \text{if } z_{k_l}+\epsilon_{k_l} > b,\\
a & \text{if } z_{k_l}+\epsilon_{k_l} < a.
\end{cases} \ \ \ \text{for } \ l=1,\dots, L.
\end{align}
Then, we construct a spectrum $\mathbf{z}\in\mathbb{R}^K$ using values $z_{k_1},\dots,z_{k_L}$ for dimensions $k_1, k_2,\dots, k_L$, respectively; while assigning zero to all the other latent dimensions. Similarly, we construct $\mathbf{\tilde{z}}\in\mathbb{R}^K$ using the perturbed values $\tilde{z}_{k_1},\dots,\tilde{z}_{k_L}$ for dimensions $k_1, k_2,\dots, k_L$, respectively, and zero for all the other dimensions. We can see that although $\mathbf{z}, \mathbf{\tilde{z}}\in\mathbb{R}^K$, they carry no information on dimensions other than $k_1, k_2,\dots, k_L$. Hence, we call a spectrum constructed in this way to be \textbf{preserved} by pattern $\mathcal{P}=\{k_1,\dots, k_L\}$. Also, we use $[a,b]_{k_l}$ to represent the domain $[a,b]$ in latent dimension $k_l$. We use $\prod_{l=1}^L [a,b]_{k_l}$ to represent the corresponding region in the latent subspace $\prod_{l=1}^L k_l$. 

Suppose $U>0$ is a given upper bound. We say that the decoder $\theta$ is \textbf{$U$-robust} with respect to pattern $\mathcal{P}=\{k_1,\dots, k_L\}$, if there exists uniform distributions $\{\mathcal{U}(-\alpha_{k_l}, \alpha_{k_l})\}_{l=1}^L$ such that, for any spectrum $\mathbf{z}\in\mathbb{R}^K$ constructed as described above from any possible values $\{a\leq z_{k_l} \leq b\}_{l=1}^L$ (i.e., any spectrum preserved by pattern $\mathcal{P}=\{k_1,\dots, k_L\}$), and for any spectrum $\mathbf{\tilde{z}}\in\mathbb{R}^K$ constructed by applying any possible perturbations $\{\epsilon_{k_l} \sim \mathcal{U}(-\alpha_{k_l}, \alpha_{k_l})\}_{l=1}^L$ to $\mathbf{z}$ (with necessary truncations as described above), we always have
\begin{align}
\|\theta(\mathbf{z})-\theta(\mathbf{\tilde{z}})\|_2\leq U.
\end{align}
Accordingly, we say that these uniform distributions $\{\mathcal{U}(-\alpha_{k_l}, \alpha_{k_l})\}_{l=1}^L$ are \textbf{$U$-qualified} with respect to pattern $\mathcal{P}=\{k_1,\dots, k_L\}$, given the decoder $\theta$. We may use `w.r.t' to simplify `with respect to'.


Intuitively, $U$-robustness means that the decoder $\theta$ is robust up to a tolerance of $U$ when we perturb a spectrum preserved by pattern $\mathcal{P}$. And trivially, if the uniform distributions $\{\mathcal{U}(-\alpha_{k_l}, \alpha_{k_l})\}_{l=1}^L$ are $U$-qualified w.r.t $\mathcal{P}=\{k_1,\dots, k_L\}$, so are $\{\mathcal{U}(-\frac{1}{2}\alpha_{k_l}, \frac{1}{2}\alpha_{k_l})\}_{l=1}^L$. Hence, when the decoder $\theta$ is $U$-robust, we can find infinitely many groups of $U$-qualified uniform distributions. Also, since $a$ and $b$ are pre-defined parameters, we will not always specify them when discussing $U$-robustness. 

Assume that the decoder $\theta$ is $U$-robust w.r.t pattern $\mathcal{P}=\{k_1,\dots, k_L\}$ given uniform distributions $\{\mathcal{U}(-\alpha_{k_l}, \alpha_{k_l})\}_{l=1}^L$. Then, we quantize the domain $[a,b]_{k_l}$ in each latent dimension $k_l\in\mathcal{P}$ by an interval size $2\alpha_{k_l}$ \citep{martinez2021permute_quantize}. Or more precisely, suppose $Q_{k_l}$ is the smallest integer larger than $\frac{b-a}{2\alpha_{k_l}}$. Then, we equally segment $[a,b]_{k_l}$ into $Q_{k_l}$ pieces. Suppose the midpoints of the segmented pieces are $q_1, q_2, \dots, q_{Q_{k_l}}$, respectively. We use these points as the quantization scales on $[a,b]_{k_l}$, and this process is applied to all the latent dimensions $k_1,\dots,k_L$ in pattern $\mathcal{P}$. Then, for any possible values $\{a\leq z_{k_l} \leq b\}_{l=1}^L$, we quantize each $z_{k_l}$ to the nearest scale on dimension $k_l$ to obtain $\{a\leq\hat{z}_{k_l}\leq b\}_{l=1}^L$. 

We can see that $|z_{k_l}-\hat{z}_{k_l}|\leq \alpha_{k_l}$ always holds true for $k_1,\dots,k_L$. In other words, $\{\hat{z}_{k_l}\}_{l=1}^L$ can be regarded as a valid perturbation of $\{ z_{k_l}\}_{l=1}^L$ based on $\{\mathcal{U}(-\alpha_{k_l}, \alpha_{k_l})\}_{l=1}^L$. Similar as described above, we construct $\mathbf{z}\in\mathbb{R}^K$ from $\{z_{k_l}\}_{l=1}^L$, and we construct $\mathbf{\hat{z}}\in\mathbb{R}^K$ from $\{\hat{z}_{k_l}\}_{l=1}^L$, with dimensions other than $k_1,\dots,k_L$ being zero. Since the decoder $\theta$ is $U$-robust w.r.t pattern $\mathcal{P}=\{k_1,\dots, k_L\}$ given uniform distributions $\{\mathcal{U}(-\alpha_{k_l}, \alpha_{k_l})\}_{l=1}^L$, we have that $\|\theta(\mathbf{z})-\theta(\mathbf{\hat{z}})\|_2\leq U$.

We can see that all the quantization scales on all latent dimensions $k_1,\dots,k_L$ can make up in total $\prod_{l=1}^LQ_{k_l}$ possible quantized sub-spectra (sub-vectors) in the region $\prod_{l=1}^L [a,b]_{k_l}$, from which we can construct $\prod_{l=1}^LQ_{k_l}$ quantized spectra on the entire latent space $\mathbb{R}^K$. We say that given the decoder $\theta$, these quantized spectra make up one \textbf{$U$-representation set} of pattern $\mathcal{P}=\{k_1,\dots, k_L\}$, denoted as $\mathcal{R}_{\mathcal{P}}$. Based on the above discussion, we can see that for any spectrum $\mathbf{z}\in\mathbb{R}^K$ preserved by pattern $\mathcal{P}$, there exists a quantized spectrum $\mathbf{\hat{z}}\in\mathcal{R}_{\mathcal{P}}$ such that $\|\theta(\mathbf{z})-\theta(\mathbf{\hat{z}})\|_2\leq U.$

As mentioned, when the decoder $\theta$ is $U$-robust w.r.t pattern $\mathcal{P}=\{k_1,\dots, k_L\}$, there exist infinitely many groups of $U$-qualified uniform distributions. Each group defines its own $U$-representation set $\mathcal{R}_{\mathcal{P}}$. But in all cases, the size of $\mathcal{R}_{\mathcal{P}}$ is always a finite integer bounded below by 1, assuming that $\theta$ is indeed $U$-robust w.r.t pattern $\mathcal{P}=\{k_1,\dots, k_L\}$. Hence, among all possible groups of $U$-qualified uniform distributions w.r.t $\mathcal{P}$, there exists a specific group $\{\mathcal{U}(-\alpha_{k_l}^*, \alpha_{k_l}^*)\}_{l=1}^L$, such that the corresponding $U$-representation set $\mathcal{R}_{\mathcal{P}}^*$ achieves the smallest possible size \citep{lohne2011vector}. 

We call $\{\mathcal{U}(-\alpha_{k_l}^*, \alpha_{k_l}^*)\}_{l=1}^L$ \textbf{$U$-optimal} with respect to pattern $\mathcal{P}=\{k_1,\dots, k_L\}$ given the decoder $\theta$. We call $\mathcal{R}_{\mathcal{P}}^*$ the \textbf{$U$-optimal representation set} of pattern $\mathcal{P}=\{k_1,\dots, k_L\}$ given the decoder $\theta$. Finally, we call the size of $\mathcal{R}_{\mathcal{P}}^*$ the \textbf{$U$-complexity} of pattern $\mathcal{P}=\{k_1,\dots, k_L\}$ with respect to the decoder $\theta$, denoted as $|\mathcal{P}|_U$. It is easy to see that with the upper bound $U$ being fixed, $|\mathcal{P}|_U$ is only determined by the latent dimensions $k_1,\dots, k_L$ in pattern $\mathcal{P}$ and the parameters in the decoder $\theta$. Also, if there do not exist uniform distributions $\{\mathcal{U}(-\alpha_{k_l}, \alpha_{k_l})\}_{l=1}^L$ such that the decoder $\theta$ is $U$-robust w.r.t pattern $\mathcal{P}=\{k_1,\dots, k_L\}$, we regard $|\mathcal{P}|_U=\infty$. 

Then, the base-2 logarithm of $U$-complexity, $\log_2(|\mathcal{P}|_U)$, measures the number of bits required to fully represent $\mathcal{R}_{\mathcal{P}}^*$: One may imagine a hash table with each key to be a binary string and its value to be a quantized spectrum $\mathbf{\hat{z}}\in\mathcal{R}_{\mathcal{P}}^*$ \citep{maurer1975hash}. This is one important concept in our theory, which will be further discussed shortly. 


\subsection{Generalization Hypotheses}\label{3.3}



The generalization ability of deep neural networks (DNNs) refers to their capacity to perform well on unseen test data after being trained on a set of training data \citep{geirhos2018generalisation}. Researchers often attribute the generalization ability of DNNs to factors such as their universal function approximation capability \citep{hanin2019universal}, continuity and smoothness of learned mappings \citep{zhou2019continuity} and the use of regularization techniques \citep{ReLU}. In this paper, without diving deeper, we merely assume that the Spectrum VAE possesses generalization ability, like other deep networks.

Generalization ability is usually assessed by the principle that a model achieving low training loss on a large training dataset is likely to achieve low test loss. In other words, as the size of the training dataset increases, a DNN with good generalization ability should become less prone to overfitting \citep{ying2019overview}. For our Spectrum VAE, we formalize this concept more rigorously as:

\begin{hypothesis}\label{hy_1}
Suppose a Spectrum VAE has an encoder parameterized by $\phi$ and a decoder parameterized by $\theta$. Given a data sample $\mathbf{x}\in\mathbb{R}^D$, suppose the encoder produces a spectrum $\mathbf{z}=\phi(\mathbf{x})\in\mathbb{R}^K$ using spiking threshold $a$ and spiking bound $b$, so that $a\leq z_k \leq b$ for $k=1,\dots, K$. Then, suppose $\mathbf{\tilde{x}}=\theta(\mathbf{z})$ is the reconstructed sample by the decoder. Let $\mathbf{x}_1, \dots, \mathbf{x}_N$ be data samples drawn from the probability distribution $\mathbf{P}$, with each $\mathbf{x}_n\in\mathbb{R}^D$. With a pre-determined upper bound $U$, we assume that for any training sample $\mathbf{x}_n\in\{\mathbf{x}_n\}_{n=1}^N$, the reconstruction error satisfies $\|\mathbf{x}_n-\mathbf{\tilde{x}}_n\|_2\leq U$, where $\mathbf{z}_n=\phi(\mathbf{x}_n)$ and $\mathbf{\tilde{x}}_n=\theta(\mathbf{z}_n)$.

\textbf{Hypothesis:} Given a new data sample $\mathbf{x}$ drawn from $\mathbf{P}$, suppose $\mathbf{z}=\phi(\mathbf{x})$ and $\mathbf{\tilde{x}}=\theta(\mathbf{z})$. Then, the larger the training sample size $N$ is, the more likely that $\|\mathbf{x}-\mathbf{\tilde{x}}\|_2\leq U$ will hold true.
\end{hypothesis}
In fact, it is trivial for us to describe Hypothesis \ref{hy_1}, since it is the fundamental principle behind deep learning \citep{goodfellow2016deep}. We simply describe Hypothesis \ref{hy_1} to make our theory complete. 

We say that a spectrum $\mathbf{z}=\phi(\mathbf{x})\in\mathbb{R}^K$ is \textbf{dormant} if it is an all-zero vector. Otherwise, we say that $\mathbf{z}$ is \textbf{active} (i.e., $z_k\geq a$ for at least one latent dimension $k$). The dormant spectrum $\mathbf{z}_0$ can only reconstruct one unique sample $\mathbf{\tilde{x}}_0=\theta(\mathbf{z}_0)$. While rare, it is possible that in Hypothesis \ref{hy_1}, a few training samples in $\{\mathbf{x}_n\}_{n=1}^N$ are reconstructed by $\mathbf{\tilde{x}}_0=\theta(\mathbf{z}_0)$ with $\|\mathbf{x}_n-\mathbf{\tilde{x}}_0\|_2 \leq U$. 


Given an active spectrum $\mathbf{z}\in\mathbb{R}^K$, suppose we observe spiking dimensions $k_1, \dots, k_L$ (i.e., we observe $z_{k_1}\geq a, \dots, z_{k_L}\geq a$). We call $\mathcal{P}=\{k_1, \dots, k_L\}$ the \textbf{spiking pattern} of $\mathbf{z}$. Based on our definition in Section \ref{3.2}, we have that $\mathbf{z}$ is preserved by $\mathcal{P}=\{k_1, \dots, k_L\}$. However, if $\mathbf{z}$ is dormant, we use $\mathcal{P}=\emptyset$ to denote its spiking pattern (we still use the term `spiking pattern'). Our second generalization hypothesis focuses exclusively on the encoder $\phi$ and spiking patterns.

Given data samples $\mathbf{x}_1,\dots,\mathbf{x}_N$, suppose the encoder $\phi$ maps them to spectra $\mathbf{z}_1,\dots,\mathbf{z}_N$ with each $\mathbf{z}_n\in\mathbb{R}^K$. From all the $N$ spectra $\{\mathbf{z}_n\}_{n=1}^N$, suppose we observe spiking patterns $\{\mathcal{P}_1,\dots,\mathcal{P}_M\}$ (with $\mathcal{P}=\emptyset$ may or may not be observed). Also, suppose each spiking pattern $\mathcal{P}_m$ is observed by $N_m$ times. It is easy to see that $\sum_{m=1}^MN_m=N$. We use $\{\mathcal{P}_1|_{N_1},\dots,\mathcal{P}_M|_{N_M}\}$ to denote the observed spiking patterns and their observed times together.

Then, we randomly select some spectra out of $\{\mathbf{z}_n\}_{n=1}^N$. From these selected spectra, we may observe some spiking patterns out of $\{\mathcal{P}_1,\dots,\mathcal{P}_M\}$. Suppose $0<P_0<1$ is a given probability. With a routine statistical analysis \citep{myers2013research}, we can prove that there exists a minimum integer $N_0>0$ such that if we randomly select $N_0$ spectra from $\{\mathbf{z}_n\}_{n=1}^N$, the probability to observe all spiking patterns $\{\mathcal{P}_1,\dots,\mathcal{P}_M\}$ from the selected spectra is at least $P_0$. We call $\frac{N}{N_0}$ the \textbf{dominant ratio} with respect to $\{\mathcal{P}_1|_{N_1},\dots,\mathcal{P}_M|_{N_M}\}$ given $P_0$, denoted as $\delta_{P_0}=\frac{N}{N_0}$.



Again, we assume the encoder has generalization ability. 
Intuitively, each spiking pattern can be viewed as a class. Then, a large dominant ratio indicates that the encoder maps a large amount of training samples to only a few classes without exception. Based on previous research \citep{zhang2016understanding, belkin2019reconciling, bengio2013representation}, this implies that the encoder genuinely generalizes different training samples into different classes rather than using a hash table approach \citep{maurer1975hash}. Hence, a new sample drawn from the same distribution will likely be assigned to existing classes—the observed spiking pattens—as well by the encoder. A simple example is provided in Appendix \ref{appen_a}. Finally, we can describe our second hypothesis: 


\begin{hypothesis}\label{hy_2}
Suppose the encoder in a Spectrum VAE is parameterized by $\phi$. Given a data sample $\mathbf{x}\in\mathbb{R}^D$, suppose the encoder produces a spectrum $\mathbf{z}=\phi(\mathbf{x})\in\mathbb{R}^K$ using spiking threshold $a$ and spiking bound $b$, so that $a\leq z_k \leq b$ for $k=1,\dots, K$. Let $\mathbf{x}_1, \dots, \mathbf{x}_N$ be data samples drawn from the probability distribution $\mathbf{P}$, with each $\mathbf{x}_n\in\mathbb{R}^D$. Suppose $\mathbf{z}_1, \dots, \mathbf{z}_N$ are the spectra generated by the encoder with $\mathbf{z}_n=\phi(\mathbf{x}_n)$ for $n=1,\dots,N$. Among $\{\mathbf{z}_1, \dots, \mathbf{z}_N\}$, suppose we observe spiking patterns $\{\mathcal{P}_1,\dots,\mathcal{P}_M\}$, with each $\mathcal{P}_m$ being observed by $N_m$ times. Finally, given $0<P_0<1$, suppose the dominant ratio w.r.t $\{\mathcal{P}_1|_{N_1},\dots,\mathcal{P}_M|_{N_M}\}$ is $\delta_{P_0}$.

\textbf{Hypothesis:} Given a new sample $\mathbf{x}$ drawn from $\mathbf{P}$, suppose $\mathbf{z}=\phi(\mathbf{x})$ is its spectrum by the encoder. Then, the larger $\delta_{P_0}$ is, the more likely that $\mathbf{z}$ is preserved by one pattern from $\{\mathcal{P}_1,\dots,\mathcal{P}_M\}$.
\end{hypothesis}


Hypothesis \ref{hy_2} claims that if the latent representations produced by a Spectrum VAE exhibit dominant spiking patterns (or `sparse combinations' of spiking latent dimensions) on the training data, this property will generalize to test data as well. With Hypotheses \ref{hy_1} and \ref{hy_2} as well as our discussion on $U$-robustness, we can describe the minimum description length of a Spectrum VAE in the next part.

\subsection{Minimum Description Length of the Spectrum VAE}\label{3.4}

The Minimum Description Length (MDL) principle states that the best model is the one that provides the shortest description of the data \citep{grunwald2007minimum_MDL_book}. Following \citep{hinton1993autoencoders_MDL}, we exclude the parameters in the encoder and decoder when calculating the description length of a Spectrum VAE, since the parameters always become negligible compared to the numerous data samples to be processed by the model after the training stage. We also exclude reconstruction errors from the description length calculation. Instead, we use mean-square-error (MSE) with an upper bound $U$ to measure information loss \citep{unruh2017information}: if reconstruction errors remain below a small $U$ for all samples drawn from the distribution $\mathbf{P}$, information loss is negligible; otherwise, a large $U$ value indicates severe information loss. Regardless, reconstruction errors are excluded when calculating the description length of a Spectrum VAE. In the rest of this paper, we may directly say `information loss bounded by $U$'. A further discussion is provided in Appendix \ref{appen_a}.

Hence, the description length of a Spectrum VAE is determined only by the spectra. To proceed, we start from describing a Spectrum VAE that is `compatible' with a given data distribution:

\begin{definition}\label{def_1}
Suppose a Spectrum VAE has an encoder parameterized by $\phi$ and a decoder parameterized by $\theta$. Given a data sample $\mathbf{x}\in\mathbb{R}^D$, suppose the encoder produces a spectrum $\mathbf{z}=\phi(\mathbf{x})\in\mathbb{R}^K$ using spiking threshold $a$ and spiking bound $b$, so that $a\leq z_k \leq b$ for $k=1,\dots, K$. Then, suppose $\mathbf{\tilde{x}}=\theta(\mathbf{z})$ is the reconstructed sample by the decoder. Let $\mathbf{x}_1, \dots, \mathbf{x}_N$ be data samples drawn from the probability distribution $\mathbf{P}$, with each $\mathbf{x}_n\in\mathbb{R}^D$. Finally, suppose $U>0$ is a given upper bound, $0<P_0<1$ is a given probability, and $\Gamma_1$, $\Gamma_2>0$ are two given thresholds. 

We say the Spectrum VAE, denoted as $(\phi,\theta)$, is \textbf{compatible} with the probability distribution $\mathbf{P}$ with respect to parameters $U$, $\Gamma_1$, $\Gamma_2$, $P_0$ and samples $\{\mathbf{x}_n\}_{n=1}^N$, if the following conditions hold:

(i) The number of data samples $N\geq \Gamma_1$. Also, for any training sample $\mathbf{x}_n$ in $\{\mathbf{x}_n\}_{n=1}^N$, the reconstruction error satisfies $\|\mathbf{x}_n-\mathbf{\tilde{x}}_n\|_2\leq U$, where $\mathbf{z}_n=\phi(\mathbf{x}_n)$ and $\mathbf{\tilde{x}}_n=\theta(\mathbf{z}_n)$;

(ii) The dominant ratio under $P_0$ with respect to $\{\mathcal{P}_1|_{N_1},\dots,\mathcal{P}_M|_{N_M}\}$ satisfies $\delta_{P_0} \geq \Gamma_2$. Here, $\{\mathcal{P}_1,\dots,\mathcal{P}_M\}$ are the spiking patterns observed from the obtained spectra $\{\mathbf{z}_1, \dots, \mathbf{z}_N\}$, and each $\mathcal{P}_m$ is observed by $N_m$ times.

We denote the set of all Spectrum VAEs that are compatible with $\mathbf{P}$ with respect to $U$, $\Gamma_1$, $\Gamma_2$, $P_0$ and $\{\mathbf{x}_n\}_{n=1}^N$ as $\mathcal{C}_{\mathbf{P}}(U, \Gamma_1,\Gamma_2,P_0,\{\mathbf{x}_n\}_{n=1}^N)$.
\end{definition}

Without specifying every-time, we assume that all Spectrum VAEs in $\mathcal{C}_{\mathbf{P}}(U, \Gamma_1,\Gamma_2,P_0,\{\mathbf{x}_n\}_{n=1}^N)$ share the same spiking threshold $a$, spiking bound $b$ and latent dimension $K$. 

Roughly speaking, compatibility requires that (i) the Spectrum VAE can reconstruct a large enough amount of training samples (ensured by a large $\Gamma_1$) drawn from the distribution $\mathbf{P}$, with reconstruction errors bounded by $U$, and (ii) the obtained spectra are preserved by a few spiking patterns that are dominant enough (ensured by a large $\Gamma_2$). If both Hypotheses \ref{hy_1} and \ref{hy_2} in Section \ref{3.3} are correct, then compatibility will imply that: Given a new data sample $\mathbf{x}$ drawn from $\mathbf{P}$, there should be a sufficiently high probability that (i) the encoder in the Spectrum VAE maps $\mathbf{x}$ to a spectrum $\mathbf{z}$ that is preserved by the observed spiking patterns, and (ii) the decoder reconstructs $\mathbf{x}$ from $\mathbf{z}$ with an error bounded by $U$.


Then, given parameters $U$, $\Gamma_1$, $\Gamma_2$, $P_0$ and training samples $\{\mathbf{x}_n\}_{n=1}^N$ drawn from the probability distribution $\mathbf{P}$, suppose we have the Spectrum VAE $(\phi,\theta)\in\mathcal{C}_{\mathbf{P}}(\frac{1}{2}U, \Gamma_1,\Gamma_2,P_0,\{\mathbf{x}_n\}_{n=1}^N)$. That is, suppose the Spectrum VAE $(\phi,\theta)$ is compatible with $\mathbf{P}$ w.r.t $\frac{1}{2}U, \Gamma_1,\Gamma_2,P_0$ and $\{\mathbf{x}_n\}_{n=1}^N$. To be specific, the upper bound for reconstruction errors is $\frac{1}{2}U$ here. Suppose we obtain the spectra $\{\mathbf{z}_n\}_{n=1}^N$ by $\mathbf{z}_n=\phi(\mathbf{x}_n)$, and suppose the observed spiking patterns (with their occurred times) are $\{\mathcal{P}_1|_{N_1},\dots,\mathcal{P}_M|_{N_M}\}$. With respect to the decoder $\theta$, suppose the $\frac{1}{2}U$-complexity (as defined in Section \ref{3.2}) of each $\mathcal{P}_m$ is $|\mathcal{P}_m|_{\frac{1}{2}U}$. Again, if the decoder $\theta$ is not $\frac{1}{2}U$-robust on pattern $\mathcal{P}_m$, then $|\mathcal{P}_m|_{\frac{1}{2}U}=\infty$. We say that the decoder $\theta$ is \textbf{$\frac{1}{2}U$-regular} with respect to the spiking patterns $\{\mathcal{P}_m\}_{m=1}^M$, if $|\mathcal{P}_m|_{\frac{1}{2}U}$ is finite for all $m=1,\dots,M$. We can naturally generalize this to obtain the definition regarding $U$-regularity, which is not repeated here.

Given a new data sample $\mathbf{x}$ drawn from $\mathbf{P}$, suppose $\mathbf{\tilde{x}}=\theta(\mathbf{z})$ and $\mathbf{z}=\phi(\mathbf{x})$. Since $(\phi,\theta)\in\mathcal{C}_{\mathbf{P}}(\frac{1}{2}U, \Gamma_1,\Gamma_2,P_0,\{\mathbf{x}_n\}_{n=1}^N)$, we are confident that $\|\mathbf{x}-\mathbf{\tilde{x}}\|_2\leq \frac{1}{2}U$, and these exists one spiking pattern $\mathcal{P}_m$ in $\{\mathcal{P}_m\}_{m=1}^M$ preserving $\mathbf{z}$. Then, suppose this pattern is $\mathcal{P}_m=\{k_{1_m},\dots,k_{L_m}\}$, and we quantize its subspace region $\prod_{l=1}^{L}[a,b]_{k_{l_m}}$ by the $\frac{1}{2}U$-optimal representation set $\mathcal{R}^*_{\mathcal{P}_m}$ (see our definition in Section \ref{3.2}). As mentioned in Section \ref{3.2}, there exists a quantized spectrum $\mathbf{\hat{z}}\in \mathcal{R}^*_{\mathcal{P}_m}$ such that $\|\theta(\mathbf{z})-\theta(\mathbf{\hat{z}})\|_2\leq \frac{1}{2}U$. Denoting $\mathbf{\hat{x}}=\theta(\mathbf{\hat{z}})$, we have 
\begin{align}\label{inequality}
\|\mathbf{x}-\mathbf{\hat{x}}\|_2\leq \|\mathbf{x}-\mathbf{\tilde{x}}\|_2+\|\mathbf{\tilde{x}}-\mathbf{\hat{x}}\|_2\leq \frac{1}{2}U+\frac{1}{2}U= U.
\end{align}
As a result, we are confident that we only need the quantized spectra in the $\frac{1}{2}U$-optimal representation sets $\mathcal{R}^*_{\mathcal{P}_1},\dots,\mathcal{R}^*_{\mathcal{P}_M}$ to reconstruct any sample $\mathbf{x}$ drawn from $\mathbf{P}$, with the reconstruction error bounded by $U$. We refer to this as the \textbf{sub-quantization trick}: When the Spectrum VAE is compatible with the distribution $\mathbf{P}$, we only need to quantize the latent subspace regions corresponding to the observed spiking patterns $\{\mathcal{P}_m\}_{m=1}^M$, rather than quantizing the entire region $\prod_{k=1}^{K}[a,b]_k$ defined by $a$ and $b$. 
This also means that to achieve satisfactory reconstructions, we are confident that we will need in total $\log_2(\sum_{m=1}^M|\mathcal{P}_m|_{\frac{1}{2}U})$ bits to transmit the primary information from the encoder to the decoder, which is the description length of the Spectrum VAE $(\phi,\theta)\in\mathcal{C}_{\mathbf{P}}(\frac{1}{2}U, \Gamma_1,\Gamma_2,P_0,\{\mathbf{x}_n\}_{n=1}^N)$.


For different Spectrum VAEs in $\mathcal{C}_{\mathbf{P}}(\frac{1}{2}U, \Gamma_1,\Gamma_2,P_0,\{\mathbf{x}_n\}_{n=1}^N)$, there can be different numbers of observed spiking patterns (i.e., different $M$) with different $\frac{1}{2}U$-complexities, leading to different values of $\sum_{m=1}^M|\mathcal{P}_m|_{\frac{1}{2}U}$. In one case, there does not exist any Spectrum VAE in $\mathcal{C}_{\mathbf{P}}(\frac{1}{2}U, \Gamma_1,\Gamma_2,P_0,\{\mathbf{x}_n\}_{n=1}^N)$ whose decoder is $\frac{1}{2}U$-regular with respect to its observed spiking patterns based on training samples $\{\mathbf{x}_n\}_{n=1}^N$. In this case, the infimum (the greatest lower bound) of $\sum_{m=1}^M|\mathcal{P}_m|_{\frac{1}{2}U}$ achieved by all Spectrum VAEs in $\mathcal{C}_{\mathbf{P}}(\frac{1}{2}U, \Gamma_1,\Gamma_2,P_0,\{\mathbf{x}_n\}_{n=1}^N)$ will be infinite, and so is the infimum of $\log_2(\sum_{m=1}^M|\mathcal{P}_m|_{\frac{1}{2}U})$ \citep{zorich2016mathematical}. 

In another case, there exists at least one Spectrum VAE in $\mathcal{C}_{\mathbf{P}}(\frac{1}{2}U, \Gamma_1,\Gamma_2,P_0,\{\mathbf{x}_n\}_{n=1}^N)$ with a $\frac{1}{2}U$-regular decoder. In this case, among all Spectrum VAEs in $\mathcal{C}_{\mathbf{P}}(\frac{1}{2}U, \Gamma_1,\Gamma_2,P_0,\{\mathbf{x}_n\}_{n=1}^N)$, there always exists a specific one achieving the minimum possible $\sum_{m=1}^M|\mathcal{P}_m|_{\frac{1}{2}U}$ \citep{lohne2011vector}, which is a finite integer. Then, the infimum of $\log_2(\sum_{m=1}^M|\mathcal{P}_m|_{\frac{1}{2}U})$ is also finite. In both cases, we have:

\begin{definition}\label{def_2}

Suppose $\mathcal{C}_{\mathbf{P}}(\frac{1}{2}U, \Gamma_1,\Gamma_2,P_0,\{\mathbf{x}_n\}_{n=1}^N)$ is the set of Spectrum VAEs that is compatible with the probability distribution $\mathbf{P}$, with respect to the given upper bound $\frac{1}{2}U$, the given probability $0<P_0<1$, the given thresholds $\Gamma_1, \Gamma_2$ and the given data samples $\{\mathbf{x}_n\}_{n=1}^N$ drawn from $\mathbf{P}$. For any Spectrum VAE $(\phi, \theta)\in\mathcal{C}_{\mathbf{P}}(\frac{1}{2}U, \Gamma_1,\Gamma_2,P_0,\{\mathbf{x}_n\}_{n=1}^N)$, suppose $\{\mathbf{z}_n\}_{n=1}^N$ are the spectra obtained from $\{\mathbf{x}_n\}_{n=1}^N$ via $\mathbf{z}_n=\phi(\mathbf{x}_n)$, and suppose $\{\mathcal{P}_1,\dots,\mathcal{P}_M\}$ are the observed spiking patterns from $\{\mathbf{z}_n\}_{n=1}^N$. Finally, for each $\mathcal{P}_m$, suppose its $\frac{1}{2}U$-complexity is $|\mathcal{P}_m|_{\frac{1}{2}U}$.

Then, with respect to the upper bound $U$, we define the \textbf{minimum description length (MDL)}, denoted as $\text{MDL}_U$, of a Spectrum VAE compatible with $\mathbf{P}$ to be the infimum of $\log_2(\sum_{m=1}^M|\mathcal{P}_m|_{\frac{1}{2}U})$ achieved by all $(\phi, \theta)\in\mathcal{C}_{\mathbf{P}}(\frac{1}{2}U, \Gamma_1,\Gamma_2,P_0,\{\mathbf{x}_n\}_{n=1}^N)$. That is,
\begin{align}\label{format_3}
\text{MDL}_U = \inf_{(\phi, \theta)\in\mathcal{C}_{\mathbf{P}}(\frac{1}{2}U, \Gamma_1,\Gamma_2,P_0,\{\mathbf{x}_n\}_{n=1}^N)} \log_2\left(\sum\nolimits_{\substack{m=1}}^M |\mathcal{P}_m|_{\frac{1}{2}U}\right).
\end{align}
Finally, if there exists $(\phi^*, \theta^*)\in\mathcal{C}_{\mathbf{P}}(\frac{1}{2}U, \Gamma_1,\Gamma_2,P_0,\{\mathbf{x}_n\}_{n=1}^N)$ achieving the $\text{MDL}_U$ based on formula \ref{format_3}, we call it the \textbf{optimal Spectrum VAE} with respect to $\mathbf{P}$, given $U$, $\Gamma_1$, $\Gamma_2$, $P_0$ and $\{\mathbf{x}_n\}_{n=1}^N$.
\end{definition}

Again, one should pay attention that, given the upper bound $U$, we describe the $\text{MDL}_U$ based on the compatibility with respect to an upper bound $\frac{1}{2}U$, and the $\frac{1}{2}U$-complexity of each spiking pattern, so that the inequality \ref{inequality} can be applied. 

Based on Formula \ref{format_3} in Definition \ref{def_2}, we can see that in order to achieve $\text{MDL}_U$, the latent representations (spectra) of a Spectrum VAE need to be robust against as large scales of perturbations as possible, so that the $\frac{1}{2}U$-complexity $|\mathcal{P}_m|_{\frac{1}{2}U}$ of each spiking pattern $\mathcal{P}_m$ can be minimized. Also, the observed spiking patterns need to be as few as possible on the training data, so that $M$ can be minimized. That is, the `combination sparsity' is required for the spiking latent dimensions. In the next part, we can explain why the optimal Spectrum VAE is `optimal'.

\subsection{Explainable Representation by Obeying MDL Principle}\label{3.5}

Suppose $\mathbf{P}$ is a data probability distribution that generates data samples $\mathbf{x}=(x_1,x_2,\dots,x_D)\in\mathbb{R}^D$. Then, we assume that the data domain is bounded in $\mathbb{R}^D$. That is, we assume that there exists boundaries $\lambda$ and $\mu$ such that $\lambda\leq x_d\leq \mu$ for each data dimension $d=1,2,\dots,D$. This is a reasonable assumption. For example, if $\mathbf{x}\in\mathbb{R}^D$ represents a flattened image vector, a typical assumption is that $0\leq x_d\leq 1$ for $d=1,\dots,D$ \citep{MNIST_paper, deng2009imagenet}.

Suppose $U>0$ is the given upper bound. By a similar analysis as in Section \ref{3.2}, we can see that we need finite many vectors in $\mathbb{R}^D$ to quantize the data domain $\prod_{d=1}^D[\lambda,\mu]$, so that for any data sample $\mathbf{x}$ drawn from $\mathbf{P}$, there exists at least one quantized vector $\mathbf{\hat{x}}$ such that $\| \mathbf{x} - \mathbf{\hat{x}}\|_2\leq U$. Moreover, we do not need to quantize the entire domain $\prod_{d=1}^D[\lambda,\mu]$. In theory, we only need to quantize the regions with a positive probability density regarding $\mathbf{P}$ \citep{renyi2007probability}. Again, among all possible quantization strategies, there exists one that requires the minimum number of quantized vectors \citep{lohne2011vector}. We call this minimum number of quantized vectors achievable by all possible quantization strategies the \textbf{$U$-essence} of the distribution $\mathbf{P}$, denoted as $\mathcal{E}_{\mathbf{P},U}$. Then, $\log_2(\mathcal{E}_{\mathbf{P},U})$ measures the number of bits required to encode the distribution $\mathbf{P}$ with information loss bounded by $U$. 

In Section \ref{3.4}, we define the compatibility of a Spectrum VAE with distribution $\mathbf{P}$ based on pre-defined parameters $U \ (\text{or} \ \frac{1}{2}U), \Gamma_1,\Gamma_2,P_0$ and obtained data samples $\{\mathbf{x}_n\}_{n=1}^N$. Given a new data sample $\mathbf{x}$ drawn from $\mathbf{P}$, sufficiently large $\Gamma_1$ and $\Gamma_2$ ensure a sufficiently high probability that (i) the encoder in the Spectrum VAE maps $\mathbf{x}$ to a spectrum $\mathbf{z}$ that is preserved by the observed spiking patterns, and (ii) the decoder reconstructs $\mathbf{x}$ from $\mathbf{z}$ with an error bounded by $U \ (\text{or} \ \frac{1}{2}U)$. But no matter how `sufficiently high' such a probability is, there is no guarantee that it equals 1. Hence, we describe an ideal scenario for theoretical analysis:

\begin{definition}\label{def_3}
Suppose $\mathbf{P}$ is a probability distribution generating data sample $\mathbf{x}\in\mathbb{R}^D$ within a bounded domain $\prod_{d=1}^D[\lambda,\mu]$. Then, given an upper bound $U>0$, we say that a Spectrum VAE $(\phi,\theta)$ is \textbf{essentially compatible} with the distribution $\mathbf{P}$ with respect to $U$, if for any sample $\mathbf{x}$ drawn from $\mathbf{P}$, we always have $\|\mathbf{x}-\mathbf{\tilde{x}}\|_2\leq U$ (again, $\mathbf{\tilde{x}}=\theta(\mathbf{z})$ and $\mathbf{z}=\phi(\mathbf{x})$). We denote the set of Spectrum VAEs essentially compatible with $\mathbf{P}$ given the upper bound $U$ to be $\mathcal{C}_{\mathbf{P},U}^\dagger$. 

For a Spectrum VAE $(\phi,\theta)\in\mathcal{C}_{\mathbf{P},\frac{1}{2}U}^\dagger$ (i.e., essentially compatible with $\mathbf{P}$ given the upper bound $\frac{1}{2}U$), suppose $\mathcal{P}_1,\dots,\mathcal{P}_M$ are all the spiking patterns we can observe on any $\mathbf{z}=\phi(\mathbf{x})$ with $\mathbf{x}$ drawn from $\mathbf{P}$ (in Appendix \ref{appen_a} we show that $M$ is finite). We use $|\mathcal{P}_m|_{\frac{1}{2}U}$ to denote the $\frac{1}{2}U$-complexity of $\mathcal{P}_m$. 

Then, with respect to the upper bound $U$, we define the \textbf{essential minimum description length (E-MDL)}, denoted as $\text{MDL}_U^\dagger$, of a Spectrum VAE essentially compatible with $\mathbf{P}$ to be the infimum of $\log_2(\sum_{m=1}^M|\mathcal{P}_m|_{\frac{1}{2}U})$ achieved by all $(\phi,\theta)\in\mathcal{C}_{\mathbf{P},\frac{1}{2}U}^\dagger$. That is,
\begin{align}\label{format_EMDL}
\text{MDL}_U^\dagger = \inf_{(\phi,\theta)\in\mathcal{C}_{\mathbf{P},\frac{1}{2}U}^\dagger} \log_2\left(\sum\nolimits_{\substack{m=1}}^M |\mathcal{P}_m|_{\frac{1}{2}U}\right).
\end{align}
Finally, if there exists $(\phi^\dagger,\theta^\dagger)\in \mathcal{C}_{\mathbf{P},\frac{1}{2}U}^\dagger$ achieving the $\text{MDL}_U^\dagger$ based on formula \ref{format_EMDL}, we call it the \textbf{essential optimal Spectrum VAE}, or \textbf{essential Spectrum VAE}, with respect to $\mathbf{P}$ given $U$.
\end{definition}

Again, each $(\phi,\theta)\in\mathcal{C}_{\mathbf{P},U}^\dagger$ (or $\mathcal{C}_{\mathbf{P},\frac{1}{2}U}^\dagger$) share the same spiking threshold $a$, spiking bound $b$ and latent dimension $K$. Intuitively, one may imagine $\Gamma_1$ and $\Gamma_2$ converging to infinity in Definition \ref{def_2} to realize essential compatibility. In theory, $\text{MDL}_U^\dagger$ measures the number of bits required by the Spectrum VAE to encode all information from the data distribution $\mathbf{P}$ with information loss bounded by $U$. Hence, $\text{MDL}_U^\dagger$ cannot be smaller than $\log_2(\mathcal{E}_{\mathbf{P},U})$, the essential minimum number of bits required to encode the same amount of information (a strict proof is provided in Appendix \ref{appen_a}). That is,

\begin{theorem}\label{thm_1}
Suppose $\mathbf{P}$ is a probability distribution generating data sample $\mathbf{x}\in\mathbb{R}^D$ within a bounded domain $\prod_{d=1}^D[\lambda,\mu]$, and suppose $U>0$ is a given upper bound. Suppose $\mathcal{C}_{\mathbf{P},\frac{1}{2}U}^\dagger$ is the set of Spectrum VAEs essentially compatible with $\mathbf{P}$ given $\frac{1}{2}U$, and suppose $\text{MDL}_U^\dagger$ is the corresponding essential minimum description length obtained by formula \ref{format_EMDL} under $\frac{1}{2}U$-complexity.

Then, we always have $\text{MDL}_U^\dagger\geq \log_2(\mathcal{E}_{\mathbf{P},U})$.
\end{theorem}

Then, for a Spectrum VAE $(\phi,\theta)\in\mathcal{C}_{\mathbf{P},\frac{1}{2}U}^\dagger$, suppose $\mathcal{P}_1,\dots,\mathcal{P}_M$ are all the spiking patterns we can observe. Suppose $\mathcal{R}^*_{\mathcal{P}_1},\dots,\mathcal{R}^*_{\mathcal{P}_M}$ are the $\frac{1}{2}U$-optimal representation set of $\mathcal{P}_1,\dots,\mathcal{P}_M$, respectively. For any quantized spectrum $\mathbf{\hat{z}}$ in any $\mathcal{R}^*_{\mathcal{P}_m}$, we say $\mathbf{\hat{z}}$ is \textbf{essentially valuable} if there exists at least one sample $\mathbf{x}$ drawn from $\mathbf{P}$ such that $\mathbf{z}=\phi(\mathbf{x})$ is quantized to $\mathbf{\hat{z}}$. All the quantized spectra within each $\mathcal{R}^*_{\mathcal{P}_m}$ that are essentially valuable further form the \textbf{$\frac{1}{2}U$-essential optimal representation set}, or \textbf{$\frac{1}{2}U$-essential representation set}, denoted as $\mathcal{R}^\dagger_{\mathcal{P}_m}$. Accordingly, we call the size of $\mathcal{R}^\dagger_{\mathcal{P}_m}$ to be the \textbf{$\frac{1}{2}U$-essential complexity} of the spiking pattern $\mathcal{P}_m$, denoted as $|\mathcal{P}_m|_{\frac{1}{2}U}^\dagger$. 

For any Spectrum VAE $(\phi,\theta)\in\mathcal{C}_{\mathbf{P},\frac{1}{2}U}^\dagger$ with spiking patterns $\{\mathcal{P}_m\}_{m=1}^M$, we call $\sum\nolimits_{\substack{m=1}}^M |\mathcal{P}_m|_{\frac{1}{2}U}-\sum\nolimits_{\substack{m=1}}^M |\mathcal{P}_m|_{\frac{1}{2}U}^\dagger$ the \textbf{$U$-residual}, which counts the unused quantized spectra in $\{\mathcal{R}^*_{\mathcal{P}_m}\}_{m=1}^M$: These quantized spectra are not useful for reconstructing data samples with errors bounded by $U$. Accordingly, we call $\sum\nolimits_{\substack{m=1}}^M |\mathcal{P}_m|_{\frac{1}{2}U}^\dagger-\mathcal{E}_{\mathbf{P},U}$ the \textbf{$U$-redundancy}, which counts the redundant quantized spectra in $\{\mathcal{R}^*_{\mathcal{P}_m}\}_{m=1}^M$: Ideally, one only needs $\mathcal{E}_{\mathbf{P},U}$ quantized vectors to represent any data sample drawn from $\mathbf{P}$ with an error bounded by $U$. Then, an ideal Spectrum VAE should also only need $\mathcal{E}_{\mathbf{P},U}$ quantized spectra for the same job. Additional quantized spectra are considered redundant. Then, we have the following theorem (whose proof is trivial, but is still provided in Appendix \ref{appen_a}): 

\begin{theorem}\label{thm_2}
Suppose $\mathbf{P}$ is a probability distribution generating data sample $\mathbf{x}\in\mathbb{R}^D$ within a bounded domain $\prod_{d=1}^D[\lambda,\mu]$, and suppose $U>0$ is a given upper bound. Suppose $\mathcal{C}_{\mathbf{P},\frac{1}{2}U}^\dagger$ is the set of Spectrum VAEs essentially compatible with $\mathbf{P}$ given $\frac{1}{2}U$, and suppose $\text{MDL}_U^\dagger$ is the corresponding essential minimum description length obtained by formula \ref{format_EMDL} under $\frac{1}{2}U$-complexity.

Then, a Spectrum VAE $(\phi^\dagger,\theta^\dagger)\in \mathcal{C}_{\mathbf{P},\frac{1}{2}U}^\dagger$ achieves $\text{MDL}_U^\dagger$ if and only if the sum of its $U$-residual and $U$-redundancy is the minimum among all Spectrum VAEs in $\mathcal{C}_{\mathbf{P},\frac{1}{2}U}^\dagger$.
\end{theorem}

\begin{figure}[htbp]
    \centering
    \includegraphics[width=1.0\textwidth]{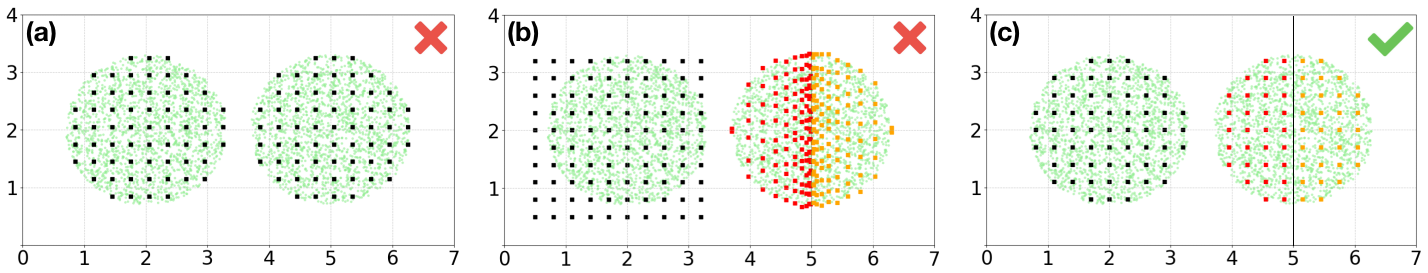}
    \caption{\underline{(Only a demo, no actual experiment)} Suppose we have a 2D uniform data distribution $\mathbf{P}$ over the union of two circles centered at $(2,2)$ and $(5,2)$, each with radius $1.2$. The solid grids in different colors represent the reconstructed samples by a Spectrum VAE $(\phi,\theta)\in\mathcal{C}_{\mathbf{P},\frac{1}{2}U}^\dagger$ using quantized spectra in different $\frac{1}{2}U$-optimal representation sets. In graph (a), reconstructed samples (black grids) from quantized spectra in the same set $\mathcal{R}^*_{\mathcal{P}_1}$ cover both circles, which is not allowed: When we perturb one quantized spectrum to its adjacent ones in $\mathcal{R}^*_{\mathcal{P}_1}$, the reconstructed samples have to stay within a distance $U$ according to the $\frac{1}{2}U$-robustness. So, grids cannot skip from one circle to another. Situation in graph (b) is not allowed for the essential Spectrum VAE $(\phi^\dagger,\theta^\dagger)$: Black grids (reconstructed samples from quantized spectra in $\mathcal{R}^*_{\mathcal{P}_1}$) exceed the left circle, leading to an increase in $U$-residual; Red and orange grids reconstructed from quantized spectra in $\mathcal{R}^*_{\mathcal{P}_2}$ and $\mathcal{R}^*_{\mathcal{P}_3}$ respectively are crowed around the boundary $x=5$, leading to an increase in $U$-redundancy. Only graph (c) is allowed for $(\phi^\dagger,\theta^\dagger)$: Quantized spectra in $\mathcal{R}^*_{\mathcal{P}_1}$ appropriately reconstruct samples in the left circle, whereas $\mathcal{R}^*_{\mathcal{P}_2}$ and $\mathcal{R}^*_{\mathcal{P}_3}$ appropriately share the right circle. Further discussions are in Appendix \ref{appen_a}.}
    \label{final_figure}
\end{figure}

Theorem \ref{thm_2} actually implies that when the $\text{MDL}_U^\dagger$ is achieved, the Spectrum VAE $(\phi^\dagger,\theta^\dagger)$ has to provide explainable latent representations: Under the grid $U$, if data samples drawn from $\mathbf{P}$ can be essentially divided into $M$ different classes, then $(\phi^\dagger,\theta^\dagger)$ has to produce at least $M$ spiking patterns, each of which should appropriately represent at most one class. A demonstration is in Figure \ref{final_figure}.

This shows another advantage of the Spectrum VAE: A small number of latent dimensions can produce a huge number of possible combinations (spiking patterns), enabling the model to represent diverse and complicated data categories in real-world. Besides, although the discussion in this sub-section is based on essential compatibility assumption, we believe that an optimal Spectrum VAE achieved on a huge training data can possess extremely similar properties.

In summary, with respect to the information loss bounded by $U$, when the minimum description length (MDL) of a Spectrum VAE is achieved, the model will automatically produce explainable latent representations of the data. That is, an explainable machine learning model, or a model with \textit{understanding} abilities, can be achieved in a self-supervised manner simply by obeying the Minimum Description Length principle. From our opinion, therefore, \textit{understanding means to represent the acquired information by as small an amount of information as possible}. Or even more briefly, \textit{to understand is to compress}.

\section{Conclusion}\label{conclusion}
In this paper, we introduced a theoretical framework for designing and evaluating deep learning architectures based on the Minimum Description Length (MDL) principle. We proposed the Spectrum VAE, whose MDL can be rigorously calculated. Also, we introduced the concept of spiking latent dimension combinations (i.e., spiking patterns) and demonstrate that minimizing the description length of a Spectrum VAE requires the observed spiking patterns to be as few as possible given enough data samples. Finally, when the MDL is achieved, the Spectrum VAE will naturally produce explainable latent representations of the data, indicating that the model \textit{understands} the data.

We hope this work can inspire future research to realize self-supervised explainable AI, or AI with understanding abilities, simply by obeying the Minimum Description Length (MDL) principle. Novel optimization algorithms other than back propagation \citep{Back_propagation} may need to be invented.

\bibliographystyle{unsrtnat}
\bibliography{references}


\appendix

\section{Supportive Analysis}\label{appen_a}

In this appendix section, we will provide supportive analysis to further evaluate our theory. We will discuss details regarding the architecture of the Spectrum VAE, provide proofs for Theorems \ref{thm_1} and \ref{thm_2}, and further discuss the examples mentioned in the main paper.\\

\textbf{Reasons to make the spiking threshold $a>0$:} If we allow the spiking threshold $a=0$, we can then `smoothly transfer' a spectrum from one spiking pattern into another by gradually reducing some latent dimensions to zero while gradually raising other latent dimensions from zero. Under $\frac{1}{2}U$-robustness, this means that the reconstructed samples will also `change smoothly' from one class to another. However, there is no guarantee that all classes of data samples drawn from $\mathbf{P}$ can be transferred smoothly in the data space $\mathbb{R}^D$ \citep{song2019generative}. This is because there is no guarantee that the domain of each data class is connected in the data space \citep{landgrebe2002hyperspectral}. Hence, allowing $a=0$ results in extra quantized spectra in different $\frac{1}{2}U$-optimal representation sets, since we need extra quantized spectra to enable this `smooth transfer' of reconstructed samples from one data domain to another disjoint data domain. In other words, allowing $a=0$ makes it impossible to really minimize the $U$-redundancy of the model. This is the reason for us to request $a>0$ when introducing the Spectrum VAE in Section \ref{3.1}.

Also, one may argue that in a Spectrum VAE, the discontinuity at the spiking threshold $a$ may jeopardize its generalization ability. We argue that this discontinuity does not significantly impair generalization: Both the encoder and the decoder in a Spectrum VAE remain to be fully differentiable networks, with the generalization capabilities intact. The encoder's mapping to preliminary latent representation $\mathbf{z}_{pre}$ and the decoder's reconstruction process should still preserve meaningful structure when trained on sufficient data. The truncation operation in formula \ref{format_1} simply enforces bounded activation on each latent dimension, without disrupting the learned manifold structure \citep{chen2024compressing} in the preliminary latent space of $\mathbf{z}_{pre}$. 

With enough training data, the encoder should learn to map similar inputs to similar preliminary latent representations, ensuring that after truncation, the spectrum is preserved by the same spiking pattern with similar value on each latent dimension. That says, the decoder will receive similar spectra if the input samples are similar. This means that the discontinuity at the spiking threshold $a$ does not cause discontinuity from the encoder to the decoder, which also means there is no discontinuity from the input to the latent representation. Therefore, we believe the discontinuity at point $a$ of each latent dimension will not significantly impact the generalization ability of the Spectrum VAE.\\

\textbf{An example on a large dominant ratio:} Suppose we have ten thousand training samples (i.e., $N=10^4$), which are mapped to ten thousand spectra in $\mathbb{R}^{16}$ (i.e., $K=16$) by the encoder. Among these spectra, suppose we observe spiking pattern $\mathcal{P}_1=\{2,3\}$ for five thousand times (i.e., $N_1=5\times 10^3$), and spiking pattern $\mathcal{P}_2=\{2,9\}$ for the other five thousand times (i.e., $N_2=5\times 10^3$). That is, we assume that dimensions $2,3$ and $9$ are the only three latent dimensions that have ever spiked across the ten thousand spectra, whereas either dimensions $2$ and $3$ spike together, or dimensions $2$ and $9$ spike together. Dimensions $3$ and $9$ are never observed spiking together. 

Then, we define $P_0=0.99$. That is, by randomly selecting $N_0$ spectra from the ten thousand ones, we want at least a $99\%$ chance to observe both $\mathcal{P}_1=\{2,3\}$ and $\mathcal{P}_2=\{2,9\}$ from the selected spectra. By a routine statistical analysis, we can get $N_0=8$ as the minimum possible numbers of selection, indicating that $\delta_{P_0}=\frac{N}{N_0}=1250$ w.r.t $\{\{2,3\}|_{5\times 10^3},\{2,9\}|_{5\times 10^3}\}$.

This simple example illustrates a situation where the dominant ratio can be considered sufficiently large. According to Hypothesis \ref{hy_2}, a new sample from the same data distribution will likely be mapped to a spectrum preserved by either $\mathcal{P}_1=\{2,3\}$ or $\mathcal{P}_2=\{2,9\}$.\\

\textbf{The reason behind saying `information loss bounded by $U$':} Suppose we have $N$ data samples $\mathbf{x}_1, \mathbf{x}_2, \dots, \mathbf{x}_N$ drawn from the probability distribution $\mathbf{P}$ with each $\mathbf{x}_n \in \mathbb{R}^D$. That is, each data sample is a vector $\mathbf{x} = (x_1, x_2, \dots, x_D)$ in a $D$-dimensional real space. Initially, we have to assume that each dimension is independent based on the Principle of Maximum Entropy \citep{guiasu1985principle}, as we lack any prior information of $\mathbf{P}$. For each dimension $d \in \{1, 2, \dots, D\}$, the data forms a length-$N$ sequence $x_{d,1}, x_{d,2}, \dots, x_{d,N}$. The differential entropy \citep{shannon1948mathematical_information_theory} for each dimension $d$, assuming a continuous distribution, is given by
$$H(X_d) = -\int p(x_d) \log p(x_d) \, dx_d,$$
where \(p(x_d)\) is the probability density function of values in dimension $d$.

However, in practice, the data values are often discrete or quantized. For instance, if $\mathbf{x}\in\mathbb{R}^D$ is a flattened image vector, then there are in total 256 possible values for each $x_d$ \cite{deng2009imagenet, MNIST_paper}. This allows us to use the discrete entropy \citep{amigo2007discrete} formula:
$$H(X_d) = -\sum_{i=1}^M p(x_d=v_i) \log p(x_d=v_i),$$
where $p(x_d=v_i)$ means the probability for $x_d$ to have value $v_i$, and $v_1,\dots,v_M$ are all the possible values $x_d$ can take. We shall say that the $M$ here has nothing to do with the number of spiking patterns in the main paper. Also, we assume that $N$ is large enough. So, the probability $p(x_d=v_i)$ is estimated based on the observed sequence $x_{d,1}, x_{d,2}, \dots, x_{d,N}$. That is, $$p(x_d=v_i)=\frac{c(x_d=v_i)}{\sum_{j=1}^Mc(x_d=v_j)},$$
where $c(x_d=v_j)$ is the number of times value $v_j$ being observed in the sequence $x_{d,1}, x_{d,2}, \dots, x_{d,N}$.

Then, summing over all $D$ dimensions, the total entropy for the entire dataset (according to the training samples) is
$$H(\mathbf{X}) = \sum_{d=1}^D H(X_d) = -\sum_{d=1}^D \sum_{i=1}^M p(x_d=v_i) \log p(x_d=v_i).$$
Now, suppose $\mathbf{\tilde{x}}_1, \mathbf{\tilde{x}}_2, \dots, \mathbf{\tilde{x}}_N$ are the reconstructed samples by a deep learning model. For each dimension $d$, we can also get a length-$N$ sequence $\tilde{x}_{d,1}, \tilde{x}_{d,2}, \dots, \tilde{x}_{d,N}$. The mutual information \citep{kraskov2004estimating} between the original and reconstructed sequences for each dimension $d$ can be calculated as
\begin{align}\label{MI}
I(X_d; \tilde{X}_d) = \sum_{i=1}^M \sum_{j=1}^M p(x_d=v_i, \tilde{x}_{d}=v_j) \log \left( \frac{p(x_d=v_i, \tilde{x}_{d}=v_j)}{p(x_d=v_i) p(\tilde{x}_{d}=v_j)} \right),
\end{align}
where \(p(x_d=v_i, \tilde{x}_{d}=v_j)\) is the joint probability of the original and reconstructed values. Similarly, assuming $N$ to be large enough, we estimate $p(x_d=v_i, \tilde{x}_{d}=v_j)$ by counting the number of times $v_i$ and $v_j$ occurring together based on the two sequences $x_{d,1}, x_{d,2}, \dots, x_{d,N}$ and $\tilde{x}_{d,1}, \tilde{x}_{d,2}, \dots, \tilde{x}_{d,N}$.

Again, summing over all dimensions, the total mutual information is
\begin{align}\label{total_MI}
I(\mathbf{X}; \mathbf{\tilde{X}}) = \sum_{d=1}^D I(X_d; \tilde{X}_d).
\end{align}
Finally, we assume that there exists an upper bound $U>0$ such that the mean-square-error $\|\mathbf{x}_n-\mathbf{\tilde{x}}_n\|_2$ is bounded by $U$ for all $n=1,\dots,N$. As discussed, the mutual information between the original and reconstructed data for each dimension is given by formula \ref{MI}, where \(p(x_d=v_i, \tilde{x}_{d}=v_j)\) represents the joint probability of the original and reconstructed values. While the mutual information formula does not directly depend on the absolute difference \(|x_{d,n} - \tilde{x}_{d,n}|\), a small bound \(U\) on the mean-square-error (MSE) indirectly influences this joint distribution:

If the MSE is bounded by a small \(U\), this implies that the dimension-wise distances are small as well. That is, we will have
$$|x_{d,n} - \tilde{x}_{d,n}| \leq U \quad \text{for all} \quad d \in \{1,2,\dots,D\},$$
This results in reconstructed samples that remain close to their original counterparts. This tight clustering around the true values in each dimension $d$ causes the joint distribution \(p(x_d=v_i, \tilde{x}_{d}=v_j)\) to concentrate along the diagonal where \(x_{d} \approx \tilde{x}_{d}\), increasing the joint probabilities for correct reconstruction.

From another point of view, mutual information can be expressed as the reduction in uncertainty about \(X_d\) after observing \(\tilde{X}_d\). That is, in terms of the entropy and conditional entropy \citep{cover1991entropy}, we have 
$$I(X_d; \tilde{X}_d) = H(X_d) - H(X_d | \tilde{X}_d).$$ 
Once again, $X_d$ is the scalar distribution of the original dimension $d$ and $\tilde{X}_d$ is that of the reconstructed dimension $d$. Then, a small \(U\) reduces the conditional entropy \(H(X_d | \tilde{X}_d)\), as the high fidelity of the reconstruction leaves less residual uncertainty, thereby increasing the overall mutual information. 

In conclusion, if the reconstruction error \(\|\mathbf{x}_n - \mathbf{\tilde{x}}_n\|_2\) is bounded by a small \(U\) for all $n=1,\dots,N$, the mutual information between the original and reconstructed values in each dimension \(d = 1, \dots, D\) will be high. This will then lead to a high total mutual information as given in formula \ref{total_MI} and, consequently, a low information loss of the reconstruction. Therefore, it is appropriate to state ``information loss bounded by \(U\)'' in the main paper.\\



\textbf{Connections between $\Gamma_1$ and $\Gamma_2$ in Definition \ref{def_1}:} As mentioned in Section \ref{3.3}, each observed spiking pattern $\mathcal{P}_m$ represents a class of data samples, which is essentially determined by the probability distribution $\mathbf{P}$, rather than by the Spectrum VAE. That says, the Spectrum VAE cannot `determine at will' the spiking pattern of a spectrum obtained from a data sample. The obtained spiking pattern has to reflect the essential class of the data sample, in order to enable a reliable reconstruction. Hence, if $\mathbf{P}$ can generate samples in a lot of different classes, then we have to collect a huge amount of data samples (obtaining a large enough $N$) if we want to achieve a large enough dominant ratio. 

This also means that increasing the threshold $\Gamma_2$ leads to increasing the required amount of training samples $N$, which in return allows a larger $\Gamma_1$. In contrast, increasing the threshold $\Gamma_1$ directly increases the required amount of training samples $N$. Since the occurring probabilities of data samples from different classes are fixed given the distribution $\mathbf{P}$, a larger $N$ will lead to a larger dominant ratio. Hence, increasing $\Gamma_1$ will allow a larger $\Gamma_2$.\\

\textbf{The number of spiking patterns $M$ in Definition \ref{def_3} is always finite:} See the proof of Theorem \ref{thm_1} in below.\\

\textbf{Proof of the two theorems in Section \ref{3.5}:}

\begin{theorem_1}
Suppose $\mathbf{P}$ is a probability distribution generating data sample $\mathbf{x}\in\mathbb{R}^D$ within a bounded domain $\prod_{d=1}^D[\lambda,\mu]$, and suppose $U>0$ is a given upper bound. Suppose $\mathcal{C}_{\mathbf{P},\frac{1}{2}U}^\dagger$ is the set of Spectrum VAEs essentially compatible with $\mathbf{P}$ given $\frac{1}{2}U$, and suppose $\text{MDL}_U^\dagger$ is the corresponding essential minimum description length obtained by formula \ref{format_EMDL} under $\frac{1}{2}U$-complexity.

Then, we always have $\text{MDL}_U^\dagger\geq \log_2(\mathcal{E}_{\mathbf{P},U})$.
\end{theorem_1}

\begin{proof}
We assume otherwise: $\text{MDL}_U^\dagger < \log_2(\mathcal{E}_{\mathbf{P},U})$. That is, there exists one Spectrum VAE $(\phi,\theta)\in\mathcal{C}_{\mathbf{P},\frac{1}{2}U}^\dagger$ who can achieve a description length lower than $\log_2(\mathcal{E}_{\mathbf{P},U})$ with reconstruction errors bounded by $U$ according to inequality \ref{inequality}. In other words, we assume that there exists one $(\phi,\theta)\in\mathcal{C}_{\mathbf{P},\frac{1}{2}U}^\dagger$ such that:

(i) $\|\mathbf{x}-\mathbf{\tilde{x}}\|_2\leq\frac{1}{2}U$ for any $\mathbf{x}$ drawn from $\mathbf{P}$, where $\mathbf{\tilde{x}}=\theta(\mathbf{z})$ and $\mathbf{z}=\phi(\mathbf{x})$;

(ii) The Spectrum VAE $(\phi,\theta)$ has spiking patterns $\mathcal{P}_1,\dots,\mathcal{P}_M$, which will preserve the spectrum $\mathbf{z}=\phi(\mathbf{x})$ given any data sample $\mathbf{x}$ drawn from $\mathbf{P}$;

(iii) $\sum\nolimits_{\substack{m=1}}^M |\mathcal{P}_m|_{\frac{1}{2}U} < \mathcal{E}_{\mathbf{P},U}$, where $|\mathcal{P}_m|_{\frac{1}{2}U}$ is the $\frac{1}{2}U$-complexity of each spiking pattern $\mathcal{P}_m$. In other words, the total amount of quantized spectra in the $\frac{1}{2}U$-optimal representation sets $\{\mathcal{R}^*_{\mathcal{P}_m}\}_{m=1}^M$ are less than $\mathcal{E}_{\mathbf{P},U}$.

Before continue, we mentioned in Definition \ref{def_3} that $M$ is always finite for any Spectrum VAE $(\phi,\theta)\in\mathcal{C}_{\mathbf{P},U}^\dagger$. This is obvious: Since the number of latent dimensions $K$ is finite, there will always be finitely many possible latent dimension combinations, though this number can be astronomical. 

Given any sample $\mathbf{x}$ drawn from $\mathbf{P}$, suppose $\mathbf{z}=\phi(\mathbf{x})$ is the spectrum, and suppose $\mathcal{P}_m$ is the spiking pattern in $\{\mathcal{P}_m\}_{m=1}^M$ that preserves $\mathbf{z}$. According to our discussion in Section \ref{3.2}, there exists a quantized spectrum $\mathbf{\hat{z}}\in\mathcal{R}^*_{\mathcal{P}_m}$ such that $\|\theta(\mathbf{z})-\theta(\mathbf{\hat{z}})\|_2\leq \frac{1}{2}U$ (in fact, $\mathbf{\hat{z}}$ is just the one which $\mathbf{z}$ quantized to in $\mathcal{R}^*_{\mathcal{P}_m}$). Denoting $\mathbf{\hat{x}}=\theta(\mathbf{\hat{z}})$, we can get $\|\mathbf{x}-\mathbf{\hat{x}}\|_2\leq U$ by the same derivation as in inequality \ref{inequality}.

This means that for any sample $\mathbf{x}$ drawn from $\mathbf{P}$, there exists one quantized vector in the set $\left\{\theta(\mathbf{\hat{z}})|\mathbf{\hat{z}}\in\{\mathcal{R}^*_{\mathcal{P}_m}\}_{m=1}^M\right\}$, such that $\|\mathbf{x}-\theta(\mathbf{\hat{z}})\|_2\leq U$. As a result, $\left\{\theta(\mathbf{\hat{z}})|\mathbf{\hat{z}}\in\{\mathcal{R}^*_{\mathcal{P}_m}\}_{m=1}^M\right\}$ is a valid quantization strategy of the data distribution $\mathbf{P}$, who contains $\sum\nolimits_{\substack{m=1}}^M |\mathcal{P}_m|_{\frac{1}{2}U}$ quantized vectors. However, we have $\sum\nolimits_{\substack{m=1}}^M |\mathcal{P}_m|_{\frac{1}{2}U} < \mathcal{E}_{\mathbf{P},U}$, which contradicts the fact that $\mathcal{E}_{\mathbf{P},U}$ is the $U$-essence (the minimum number of quantized vectors achievable by any quantization strategy) of $\mathbf{P}$. As a result, our original assumption is wrong. Hence, $\text{MDL}_U^\dagger\geq \log_2(\mathcal{E}_{\mathbf{P},U})$ always holds true.
\end{proof}

\begin{theorem_2}
Suppose $\mathbf{P}$ is a probability distribution generating data sample $\mathbf{x}\in\mathbb{R}^D$ within a bounded domain $\prod_{d=1}^D[\lambda,\mu]$, and suppose $U>0$ is a given upper bound. Suppose $\mathcal{C}_{\mathbf{P},\frac{1}{2}U}^\dagger$ is the set of Spectrum VAEs essentially compatible with $\mathbf{P}$ given $\frac{1}{2}U$, and suppose $\text{MDL}_U^\dagger$ is the corresponding essential minimum description length obtained by formula \ref{format_EMDL} under $\frac{1}{2}U$-complexity.

Then, a Spectrum VAE $(\phi^\dagger,\theta^\dagger)\in \mathcal{C}_{\mathbf{P},\frac{1}{2}U}^\dagger$ achieves $\text{MDL}_U^\dagger$ if and only if the sum of its $U$-residual and $U$-redundancy is the minimum among all Spectrum VAEs in $\mathcal{C}_{\mathbf{P},\frac{1}{2}U}^\dagger$.
\end{theorem_2}

\begin{proof}
Suppose a Spectrum VAE $(\phi^\dagger,\theta^\dagger)\in \mathcal{C}_{\mathbf{P},\frac{1}{2}U}^\dagger$ achieves the essential minimum description length $\text{MDL}_U^\dagger$ by formula \ref{format_EMDL} under $\frac{1}{2}U$-complexity. That is, $(\phi^\dagger,\theta^\dagger)$ achieves the minimum $\sum\nolimits_{\substack{m=1}}^M |\mathcal{P}_m|_{\frac{1}{2}U}$ among all Spectrum VAEs in $\mathcal{C}_{\mathbf{P},\frac{1}{2}U}^\dagger$, where $\mathcal{P}_1,\dots,\mathcal{P}_M$ are all the spiking patterns produced by the encoder $\phi^\dagger$ that we can observe. Then, since $\mathcal{E}_{\mathbf{P},U}$ is independent of any Spectrum VAE, we have that $(\phi^\dagger,\theta^\dagger)$ also achieves the minimum
\begin{align}\label{formula_proof_thm_2}
&\sum\nolimits_{\substack{m=1}}^M |\mathcal{P}_m|_{\frac{1}{2}U}-\mathcal{E}_{\mathbf{P},U}\notag\\
=&\sum\nolimits_{\substack{m=1}}^M |\mathcal{P}_m|_{\frac{1}{2}U}-\sum\nolimits_{\substack{m=1}}^M |\mathcal{P}_m|_{\frac{1}{2}U}^\dagger + \sum\nolimits_{\substack{m=1}}^M |\mathcal{P}_m|_{\frac{1}{2}U}^\dagger-\mathcal{E}_{\mathbf{P},U}
\end{align}
among all Spectrum VAEs in $\mathcal{C}_{\mathbf{P},\frac{1}{2}U}^\dagger$, which is the sum of $U$-residual and $U$-redundancy.

Conversely, suppose a Spectrum VAE $(\phi^\dagger,\theta^\dagger)\in \mathcal{C}_{\mathbf{P},\frac{1}{2}U}^\dagger$ achieves the minimum sum of $U$-residual and $U$-redundancy among all Spectrum VAEs in $\mathcal{C}_{\mathbf{P},\frac{1}{2}U}^\dagger$. According to formula \ref{formula_proof_thm_2}, $(\phi^\dagger,\theta^\dagger)$ also achieves the minimum $\sum\nolimits_{\substack{m=1}}^M |\mathcal{P}_m|_{\frac{1}{2}U}-\mathcal{E}_{\mathbf{P},U}$ among all Spectrum VAEs in $\mathcal{C}_{\mathbf{P},\frac{1}{2}U}^\dagger$. Since $\mathcal{E}_{\mathbf{P},U}$ is independent of any Spectrum VAE, we have that $(\phi^\dagger,\theta^\dagger)$ also achieves the minimum $\sum\nolimits_{\substack{m=1}}^M |\mathcal{P}_m|_{\frac{1}{2}U}$, and hence the minimum $\log_2(\sum\nolimits_{\substack{m=1}}^M |\mathcal{P}_m|_{\frac{1}{2}U})$, among all Spectrum VAEs in $\mathcal{C}_{\mathbf{P},\frac{1}{2}U}^\dagger$. Therefore, $(\phi^\dagger,\theta^\dagger)$ achieves $\text{MDL}_U^\dagger$ according to formula \ref{format_EMDL}, in which case the minimum $\log_2(\sum\nolimits_{\substack{m=1}}^M |\mathcal{P}_m|_{\frac{1}{2}U})$ exists and hence the infimum equals the minimum.
\end{proof}

\textbf{Further discussion on Figure \ref{final_figure}}

Once again, we have to mention that all graphs in Figure \ref{final_figure} are made only for demonstration purpose. There is no actual experimental result involved. Suppose we have a 2D uniform data distribution $\mathbf{P}$ over the union of two circles centered at $(2,2)$ and $(5,2)$, each with radius $1.2$.

\begin{figure}[htbp]
    \centering
    \includegraphics[width=0.45\textwidth]{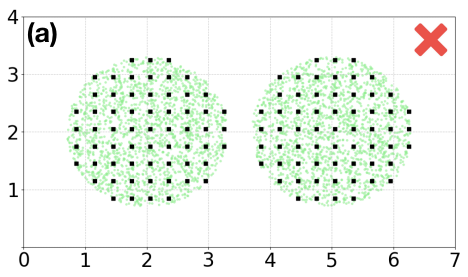}
    \caption*{Graph (a): Reconstructed samples from quantized spectra in the same $\frac{1}{2}U$-optimal representation set cover both circles. This is impossible: forbidden by $\frac{1}{2}U$-robustness.}
    \label{graph_a}
\end{figure}

\textbf{Discussing graph (a):} Graph (a) describes that given a Spectrum VAE $(\phi,\theta) \in \mathcal{C}_{\mathbf{P},\frac{1}{2}U}^\dagger$, the reconstructed samples (black grids) from quantized spectra in the same $\frac{1}{2}U$-optimal representation set $\mathcal{R}^*_{\mathcal{P}_1}$ cover both circles. However, this is actually impossible: 

As we discussed in Section \ref{3.2}, when the decoder $\theta$ is $\frac{1}{2}U$-robust with respect to the spiking pattern $\mathcal{P}_1=\{k_1,\dots,k_L\}$, there exists a specific group of uniform distributions $\{\mathcal{U}(-\alpha_{k_l}^*, \alpha_{k_l}^*)\}_{l=1}^L$, such that the corresponding $\frac{1}{2}U$-representation set $\mathcal{R}^*_{\mathcal{P}_1}$ achieves the smallest possible size, which is the optimal one. Also, remember that $\mathcal{R}^*_{\mathcal{P}_1}$ is obtained by quantizing each latent dimension $k_l$ using the interval size $2\alpha_{k_l}^*$. Or more precisely, with $Q_{k_l}^*$ to be the smallest integer larger than $(b-a)/(2\alpha_{k_l}^*)$, we equally divide the interval $[a,b]_{k_l}$ into $Q_{k_l}^*$ segments and take the middle point of each segment as the quantization scale. Finally, we construct a quantized spectrum $\mathbf{\hat{z}}\in\mathbb{R}^K$ from any possible quantization scales on dimension $k_1,\dots,k_L$, and other dimensions being zero. All possible $\mathbf{\hat{z}}$ make up $\mathcal{R}^*_{\mathcal{P}_1}$.

Then, suppose $\mathbf{\hat{z}}=(\hat{z}_1,\dots,\hat{z}_K)$ is a quantized spectra in $\mathcal{R}^*_{\mathcal{P}_1}$. According to Section \ref{3.2}, $\mathbf{\hat{z}}$ is preserved by pattern $\mathcal{P}_1=\{k_1,\dots,k_L\}$. That is, $\hat{z}_k=0$ if $k\notin \{k_1,\dots,k_L\}$. Also, for $l=1,\dots,L$, $a\leq \hat{z}_{k_l}\leq b$ is one quantization scale on dimension $k_l$. Suppose we choose a specific latent dimension $k_l$ in $\{k_1,\dots,k_L\}$ and perturb $\hat{z}_{k_l}$ to $\hat{z}_{k_l}+\alpha_{k_l}^*$, which provide us $\mathbf{\tilde{z}}=(\hat{z}_1,\dots,\hat{z}_{k_l}+\alpha_{k_l}^*,\dots,\hat{z}_K)$. Based on the definition of $\frac{1}{2}U$-robustness, we have that $\|\theta(\mathbf{\tilde{z}})-\theta(\mathbf{\hat{z}})\|_2\leq \frac{1}{2}U$. Then, we carry on perturbing $\hat{z}_{k_l}+\alpha_{k_l}^*$ to $\hat{z}_{k_l}'$, the adjacent quantization scale of $\hat{z}_{k_l}$, which provide us $\mathbf{\hat{z}}'=(\hat{z}_1,\dots,\hat{z}_{k_l}',\dots,\hat{z}_K)$. Once again, based on $\frac{1}{2}U$-robustness, we have $\|\theta(\mathbf{\hat{z}}')-\theta(\mathbf{\tilde{z}})\|_2\leq \frac{1}{2}U$. As a result, we have
\begin{align}\label{inequal_2}
\|\theta(\mathbf{\hat{z}}')-\theta(\mathbf{\hat{z}})\|_2\leq \|\theta(\mathbf{\hat{z}}')-\theta(\mathbf{\tilde{z}})\|_2 + \|\theta(\mathbf{\tilde{z}})-\theta(\mathbf{\hat{z}})\|_2\leq \frac{1}{2}U+\frac{1}{2}U=U.
\end{align}
In fact, based on the definition of $\frac{1}{2}U$-robustness, we can perturb multiple latent dimensions in $\{k_1,\dots,k_L\}$ of $\mathbf{\hat{z}}$ simultaneously to the adjacent quantization scales, and still have inequality \ref{inequal_2} holding true. But in order to produce a straightforward analysis, we only pick one specific dimension $k_l$ here for perturbation.

Aa a result, we can conclude that when we perturb one quantized spectrum $\mathbf{\hat{z}}$ to its adjacent ones in $\mathcal{R}^*_{\mathcal{P}_1}$, the reconstructed samples have to stay within a distance $U$ according to the $\frac{1}{2}U$-robustness. This is also what we mentioned in Section \ref{3.5} when introducing graph (a). In practice, $U$ should be small enough, so that the reconstructed samples after perturbing $\mathbf{\hat{z}}\in\mathcal{R}^*_{\mathcal{P}_1}$ cannot skip from one data domain to another disjoint data domain, like the two disjoint circles in our example. This tells us, intuitively, the reconstructed samples from all the quantized spectra in one $\frac{1}{2}U$-optimal representation set cannot scatter across the data space. Rather, reconstructed samples from adjacent quantized spectra have to stay within a distance $U$ of each other, like molecules in crystal.

By a similar analysis, we can conclude that reconstructed samples from all spectra preserved by one spiking pattern cannot scatter across the data space, either. This property forbids one spiking pattern to represent data samples from multiple classes, which is important in our theory.\\

\textbf{Discussing graph (b):} Graph (b) describes a Spectrum VAE $(\phi,\theta) \in \mathcal{C}_{\mathbf{P},\frac{1}{2}U}^\dagger$ which cannot be the essential Spectrum VAE.

\begin{figure}[htbp]
    \centering
    \includegraphics[width=0.45\textwidth]{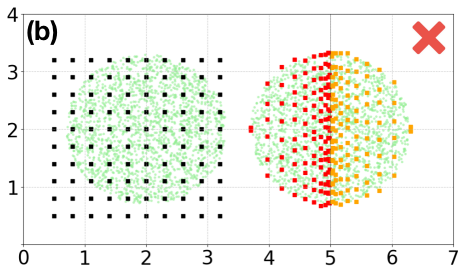}
    \caption*{Graph (b): Black grids (reconstructed samples from quantized spectra in one $\frac{1}{2}U$-optimal representation set) exceed the left circle, leading to an increase in $U$-residual; Red and orange grids reconstructed from quantized spectra in two different $\frac{1}{2}U$-optimal representation sets are crowded around the boundary $x=5$, leading to an increase in $U$-redundancy.}
    \label{graph_b}
\end{figure}

\underline{First, we discuss the black grids in graph (b)}: The black grids represent the reconstructed samples by the decoder $\theta$ from quantized spectra in the $\frac{1}{2}U$-optimal representation set $\mathcal{R}^*_{\mathcal{P}_1}$. We can see that there are some reconstructed samples beyond the left circle. Then, suppose the quantized spectrum $\mathbf{\hat{z}}\in\mathcal{R}^*_{\mathcal{P}_1}$ reconstructs such a sample beyond the left circle (i.e., $\mathbf{\hat{x}}=\theta(\mathbf{\hat{z}})$ is beyond the left circle). Then, we assume that there exists one sample $\mathbf{x}$ drawn from the data distribution $\mathbf{P}$ such that $\mathbf{z}=\phi(\mathbf{x})$ is quantized to $\mathbf{\hat{z}}$. Since $(\phi,\theta) \in \mathcal{C}_{\mathbf{P},\frac{1}{2}U}^\dagger$, the reconstruction error for $\mathbf{x}$ will be bounded by $\frac{1}{2}U$. That is, we have $\|\mathbf{x}-\mathbf{\tilde{x}}\|_2\leq \frac{1}{2}U$, where $\mathbf{\tilde{x}}=\theta(\mathbf{z})$. It is easy to see that $\mathbf{x}$ is within the left circle, since $\mathbf{x}$ is drawn from $\mathbf{P}$. Hence, $\mathbf{\tilde{x}}=\theta(\mathbf{z})$ will be within or very close to the left circle.

On the other hand, we have $\mathbf{\hat{x}}=\theta(\mathbf{\hat{z}})$ is beyond the left circle, and $\mathbf{z}$ is quantized to $\mathbf{\hat{z}}$. That is, when perturbing $\mathbf{z}$ to $\mathbf{\hat{z}}$, the reconstructed sample skips from $\theta(\mathbf{z})$ to $\theta(\mathbf{\hat{z}})$, which is from within the left circle to beyond the left circle. Since $\frac{1}{2}U$ is small enough in practice, $\|\theta(\mathbf{z})-\theta(\mathbf{\hat{z}})\|_2$ will be larger than $\frac{1}{2}U$, which violates the $\frac{1}{2}U$-robustness. As a result, our assumption at beginning is wrong. That says, if a quantized spectrum $\mathbf{\hat{z}}\in\mathcal{R}^*_{\mathcal{P}_1}$ reconstructs the sample $\mathbf{\hat{x}}=\theta(\mathbf{\hat{z}})$ beyond the left circle, there does not exist any data sample $\mathbf{x}$ drawn from $\mathbf{P}$ such that $\mathbf{z}=\phi(\mathbf{x})$ is quantized to $\mathbf{\hat{z}}$. As a result, such a quantized spectrum $\mathbf{\hat{z}}\in\mathcal{R}^*_{\mathcal{P}_1}$ cannot be essentially valuable, indicating that $\mathbf{\hat{z}}\notin\mathcal{R}^\dagger_{\mathcal{P}_1}$. 

Therefore, if some quantized spectra in one $\frac{1}{2}U$-optimal representation set reconstruct samples beyond the data domain of the probability distribution, then the $U$-residual of the Spectrum VAE will increase. This implies that for the essential Spectrum VAE $(\phi^\dagger,\theta^\dagger)\in \mathcal{C}_{\mathbf{P},\frac{1}{2}U}^\dagger$, the quantized spectrum in any of its $\frac{1}{2}U$-optimal representation set must reconstruct the sample within the domain where the data distribution $\mathbf{P}$ has a positive probability density. Intuitively, $(\phi^\dagger,\theta^\dagger)$ cannot waste its spectrum to represent a data sample that will never be generated by $\mathbf{P}$.\\

\underline{Then, we discuss the red and orange grids in graph (b):} The red and orange grids represent the reconstructed samples by the decoder $\theta$ from quantized spectra in two different $\frac{1}{2}U$-optimal representation sets, $\mathcal{R}^*_{\mathcal{P}_2}$ and $\mathcal{R}^*_{\mathcal{P}_3}$ respectively. That is, quantized spectra in $\mathcal{R}^*_{\mathcal{P}_2}$ reconstruct red grids whereas quantized spectra in $\mathcal{R}^*_{\mathcal{P}_3}$ reconstruct orange grids. The red and orange grids share the right circle. Also, suppose $\mathcal{P}_2$ and $\mathcal{P}_3$ are the spiking pattens regarding $\mathcal{R}^*_{\mathcal{P}_2}$ and $\mathcal{R}^*_{\mathcal{P}_3}$, respectively.

In our example, any data sample $\mathbf{x}$ in the right circle will be reconstructed by the Spectrum VAE $(\phi,\theta)$ with an error bounded by $\frac{1}{2}U$, since $(\phi,\theta) \in \mathcal{C}_{\mathbf{P},\frac{1}{2}U}^\dagger$. To be specific, according to our assumption, for any data sample $\mathbf{x}$ in the right circle, $\mathbf{z}=\phi(\mathbf{x})$ is preserved by either $\mathcal{P}_2$ or $\mathcal{P}_3$, without exception. Hence, we use $\mathcal{D}_{\mathcal{P}_2}$ to denote the 2D domain in the right circle consisting of data samples $\mathbf{x}$ such that $\mathbf{z}=\phi(\mathbf{x})$ is preserved by $\mathcal{P}_2$. That is, $\mathcal{D}_{\mathcal{P}_2}=\{\mathbf{x}|\mathbf{x} \ \text{is drawn from} \ \mathbf{P} \ \text{and} \ \mathbf{z}=\phi(\mathbf{x}) \ \text{is preserved by} \ \mathcal{P}_2\}$. Similarly, we define $\mathcal{D}_{\mathcal{P}_3}=\{\mathbf{x}|\mathbf{x} \ \text{is drawn from} \ \mathbf{P} \ \text{and} \ \mathbf{z}=\phi(\mathbf{x}) \ \text{is preserved by} \ \mathcal{P}_3\}$. It is easy to see that $\mathcal{D}_{\mathcal{P}_2}\cap \mathcal{D}_{\mathcal{P}_3}=\emptyset$ and $\mathcal{D}_{\mathcal{P}_2}\cup \mathcal{D}_{\mathcal{P}_3}$ is the right circle.

Then, there must be boundaries within the right circle between $\mathcal{D}_{\mathcal{P}_2}$ and $\mathcal{D}_{\mathcal{P}_3}$. In graph (b), we simply assume that there is only one boundary, $x=5$, for straightforward demonstration purpose. Before continue, we note that in this example, we have 2D data samples. That is, we have $\mathbf{x}=(x,y)\in\mathbb{R}^2$. One should differ a data sample $\mathbf{x}$ (in bold format) from the boundary $x=5$ (in regular format), although both of them use the symbol `$x$'. 

One can see in graph (b) that the red and orange grids are crowed together around the boundary $x=5$. Through this phenomenon, we want to demonstrate that the reconstructed samples from quantized spectra in $\mathcal{R}^*_{\mathcal{P}_2}$ are closer to each other than $U$, if these reconstructed samples approach $x=5$ from the left side. Similarly, when the reconstructed samples from quantized spectra in $\mathcal{R}^*_{\mathcal{P}_3}$ approach $x=5$ from the right side, these reconstructed samples are closer to each other than $U$. 

Then, for a data sample $\mathbf{x}$ approaching $x=5$ from the left side, there can be multiple reconstructed samples from quantized spectra in $\mathcal{R}^*_{\mathcal{P}_2}$ such that these reconstructed samples are closer to $\mathbf{x}$ than $U$. That is, $\mathbf{x}$ can be quantized to multiple spectra in $\mathcal{R}^*_{\mathcal{P}_2}$ to achieve a reconstruction with an error bounded by $U$. So, by carefully selecting the quantized spectra for all possible data samples, we can successfully avoid using some quantized spectra in $\mathcal{R}^*_{\mathcal{P}_2}$ at all. As discussed in Section \ref{3.5}, an ideal Spectrum VAE should only need $\mathcal{E}_{\mathbf{P},U}$ quantized spectra to represent any data sample drawn from $\mathbf{P}$ with an error bounded by $U$. Then, these quantized spectra being avoided in $\mathcal{R}^*_{\mathcal{P}_2}$ become redundant, leading to an increase in the $U$-redundancy. Exactly the same analysis can applied to $\mathcal{R}^*_{\mathcal{P}_3}$.

Intuitively, for the essential Spectrum VAE $(\phi^\dagger,\theta^\dagger)$, the reconstructed samples from the quantized spectra in each $\frac{1}{2}U$-optimal representation set cannot be crowed to each other. Otherwise there will be redundant quantized spectra carrying similar information, which ultimately increase the $U$-redundancy $\sum\nolimits_{\substack{m=1}}^M |\mathcal{P}_m|_{\frac{1}{2}U}^\dagger-\mathcal{E}_{\mathbf{P},U}$ of the model.


One may argue that since we choose the $\frac{1}{2}U$-optimal representation set for each spiking pattern, we will not observe crowed grids (again, `grids' refers to the reconstructed samples from each $\frac{1}{2}U$-optimal representation set). Or in other words, one may say there will be no redundant quantized spectra in an $\frac{1}{2}U$-optimal representation set, since it is already the optimal representation set given the upper bound $\frac{1}{2}U$. But this is not true: 

Given a Spectrum VAE $(\phi,\theta)$, suppose the decoder $\theta$ is $\frac{1}{2}U$-robust with respect to pattern $\mathcal{P}=\{k_1,\dots,k_L\}$. As we discussed in Section \ref{3.2}, suppose the group of uniform distributions $\{\mathcal{U}(-\alpha_{k_l}^*, \alpha_{k_l}^*)\}_{l=1}^L$ is $\frac{1}{2}U$-optimal with respect to $\mathcal{P}=\{k_1,\dots,k_L\}$, and the corresponding set of quantized spectra $\mathcal{R}^*_{\mathcal{P}}$ is the $\frac{1}{2}U$-optimal representation set. According to Section \ref{3.2}, this means that the size of $\mathcal{R}^*_{\mathcal{P}}$ is the minimum among all $\frac{1}{2}U$-representation sets. Then, suppose $\{\mathcal{U}(-\alpha_{k_l}', \alpha_{k_l}')\}_{l=1}^L$ is another group of uniform distributions, where $\alpha_{k_l}'\geq \alpha_{k_l}^*$ hold true for all $l=1,\dots,L$, and the strict `$>$' relationship is true for at least one $l$. If we quantize latent dimensions in $\{k_1,\dots,k_L\}$ based on $\{\mathcal{U}(-\alpha_{k_l}', \alpha_{k_l}')\}_{l=1}^L$ and obtain the set of quantized spectra $\mathcal{R}'_{\mathcal{P}}$, we can see that the size of $\mathcal{R}'_{\mathcal{P}}$ is smaller than that of $\mathcal{R}^*_{\mathcal{P}}$.

But the size of $\mathcal{R}^*_{\mathcal{P}}$ is already the minimum among all $\frac{1}{2}U$-representation sets. So, we know that this group of uniform distributions $\{\mathcal{U}(-\alpha_{k_l}', \alpha_{k_l}')\}_{l=1}^L$ cannot guarantee the $\frac{1}{2}U$-robustness of decoder $\theta$ with respect to $\mathcal{P}=\{k_1,\dots,k_L\}$. That is, there exists at least one spectrum $\mathbf{z}'$ preserved by $\mathcal{P}$ and one specific perturbation $\{\epsilon_{k_l}' \sim \mathcal{U}(-\alpha_{k_l}', \alpha_{k_l}')\}_{l=1}^L$, such that if we perturb $\mathbf{z}'$ to $\mathbf{z}''$ using $\{\epsilon_{k_l}'\}_{l=1}^L$ (with necessary truncation back to $[a,b]$ on each latent dimension $k_l$ as described in Section \ref{3.2}), then we will have $\|\theta(\mathbf{z}')-\theta(\mathbf{z}'')\|_2>\frac{1}{2}U$.

We can view such a spectrum $\mathbf{z}'$ as the sensitive point for the decoder $\theta$. However, it is not necessary for all spectra preserved by $\mathcal{P}=\{k_1,\dots,k_L\}$ to be as sensitive. That is, intuitively, the perturbation ranges $\{\alpha_{k_l}^*\}_{l=1}^L$ in the $\frac{1}{2}U$-optimal $\{\mathcal{U}(-\alpha_{k_l}^*, \alpha_{k_l}^*)\}_{l=1}^L$ is decided by the most sensitive points in the latent space preserved by $\mathcal{P}=\{k_1,\dots,k_L\}$. It is totally possible that there exist other less sensitive regions in the latent space preserved by $\mathcal{P}$, where spectra can be perturbed by much larger scales while still keep the reconstruction differences being bounded by $\frac{1}{2}U$. Therefore, adjacent quantized spectra in $\mathcal{R}^*_{\mathcal{P}}$ within these less sensitive regions will carry similar information, and hence be redundant. This is because when we perturb one spectrum to its adjacent ones in these less sensitive regions, the reconstruction differences can be much less than $\frac{1}{2}U$. This leads to an increase in the $U$-redundancy and causes crowed grids as we demonstrated.

Therefore, in order to reduce the $U$-redundancy, parameters in the Spectrum VAE should guarantee `equal sensitivity' against perturbations as much as possible among all spectra preserved by each spiking pattern. Anyway, merely applying $\frac{1}{2}U$-optimal representation sets (or applying $\frac{1}{2}U$-optimal uniform distributions) will not necessarily minimize the $U$-redundancy.\\

\textbf{Discussing graph (c):} Based on the above discussion, we learned that: 

\begin{figure}[htbp]
    \centering
    \includegraphics[width=0.45\textwidth]{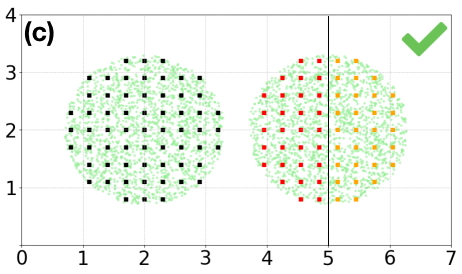}
    \caption*{Graph (c): Quantized spectra in $\mathcal{R}^*_{\mathcal{P}_1}$ appropriately represent samples in the left circle, whereas $\mathcal{R}^*_{\mathcal{P}_2}$ and $\mathcal{R}^*_{\mathcal{P}_3}$ appropriately share the right circle. This allows the $U$-residual and $U$-redundancy to be minimized.}
    \label{graph_c}
\end{figure}

(i) Due to $\frac{1}{2}U$-robustness, reconstructed samples from quantized spectra in one $\frac{1}{2}U$-optimal representation set cannot skip from one data domain into another disjoint data domain. Due to the same reason, reconstructed samples from general spectra preserved by one spiking pattern cannot scatter across the data space, either.

(ii) Reconstructed samples from quantized spectra in one $\frac{1}{2}U$-optimal representation set cannot exceed the data domain where the data distribution $\mathbf{P}$ has a positive probability density. Otherwise the quantized spectra are wasted to reconstruct never existed samples, leading to an increase in the $U$-redundancy.

(iii) Reconstructed samples from quantized spectra in one $\frac{1}{2}U$-optimal representation set cannot become crowed together with distances to each other smaller than $U$. Otherwise some quantized spectra in the $\frac{1}{2}U$-optimal representation set will carry similar information, leading to an increase in the $U$-redundancy. A necessary condition to avoid crowed reconstructions is that all spectra preserved by the spiking pattern (not only the quantized ones in the $\frac{1}{2}U$-optimal representation set) are `equally sensitive' to random perturbations.

Therefore, only the situation described in graph (c) is allowed for an essential Spectrum VAE $(\phi^\dagger,\theta^\dagger)$: There are three spiking patterns, $\mathcal{P}_1, \mathcal{P}_2$ and $\mathcal{P}_3$, which are all we can observe from any spectrum. Then, suppose $\mathcal{R}^*_{\mathcal{P}_1}, \mathcal{R}^*_{\mathcal{P}_2}$ and $\mathcal{R}^*_{\mathcal{P}_3}$ are their $\frac{1}{2}U$-optimal representation sets, respectively. We can see in the graph that reconstructed samples from quantized spectra in each $\frac{1}{2}U$-optimal representation set stay within one circle rather than scatter around. Also, no quantized spectrum in $\mathcal{R}^*_{\mathcal{P}_1}, \mathcal{R}^*_{\mathcal{P}_2}$ and $\mathcal{R}^*_{\mathcal{P}_3}$ is used to reconstruct any sample exceeding the two circles, indicating that residual quantized spectra are minimized. Finally, crowed grids are not observed, which is necessary for minimizing redundant quantized spectra.\\

\textbf{Further discussing graph (c):} One may argue that what we demonstrate in graph (c) is still not the optimal situation. In an optimal situation, each spiking pattern of the essential Spectrum VAE should represent one entire class of data. In other words, each class of data should be represented by exactly one spiking pattern. Multiple spiking patterns will not share any data class for representation, as described in graph (d):

\begin{figure}[htbp]
    \centering
    \includegraphics[width=0.45\textwidth]{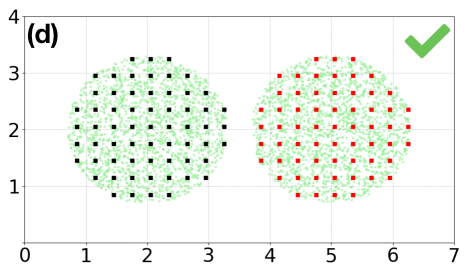}
    \caption*{Graph (d): Quantized spectra in $\mathcal{R}^*_{\mathcal{P}_1}$ appropriately reconstruct samples in the left circle, whereas quantized spectra in $\mathcal{R}^*_{\mathcal{P}_2}$ appropriately reconstruct samples in the right circle. Spiking patterns $\mathcal{P}_1$ and $\mathcal{P}_2$ do not share the same data class.}
    \label{graph_d}
\end{figure}

We admit that such an optimal situation is intuitively more perfect. In addition, we provide our intuitive analysis to support such an optimal situation: 

If multiple spiking patterns share one data class, based on the above discussion, the reconstructed samples from spectra preserved by different spiking patterns will have boundaries between each other. On the other hand, the essential Spectrum VAE needs to minimize the residual and redundant quantized spectra in its $\frac{1}{2}U$-optimal representation sets given the upper bound $U$. This means that the reconstructed samples from the quantized spectra in each $\frac{1}{2}U$-optimal representation set cannot exceed the boundaries, and cannot become crowed around the boundaries or within the data domain divided by the boundaries. 

This implies that the information of the existence of the boundaries are captured by the latent representation already: Quantized spectra in each $\frac{1}{2}U$-optimal representation set needs to `know' where the boundaries are, so that the reconstructed samples will not exceed the boundaries and will not become crowded near the boundaries. Such information of the boundaries is not only captured in the parameters of the encoder and decoder, but also needs to be preserved by the latent representations.

However, the boundaries has nothing to do with the data probability distribution. As a result, the information of the boundaries is redundant. Hence, preserving the information of the boundaries in the latent representations will increase the $U$-redundancy of each $\frac{1}{2}U$-optimal representation set. This implies that the spiking patterns of the essential Spectrum VAE need to minimize the boundaries among their reconstructed samples. So, situation in graph (c) is not fully optimized, which is not for the essential Spectrum VAE.

Note that this intuitive analysis is lack of mathematical description, definition and derivation. As a result, we only include it in the Appendix. In our main theory, we still claim this: Under the grid $U$, if data samples drawn from $\mathbf{P}$ can be essentially divided into $M$ different classes, then the essential Spectrum VAE $(\phi^\dagger,\theta^\dagger)$ has to produce at least $M$ spiking patterns, each of which should appropriately represent at most one class. Sharing the same data class by multiple spiking patterns for representation is not in theory forbidden, as demonstrated in the right circle of graph (c). 

Recall that we defined in Section \ref{3.5} the $U$-essence of a data probability distribution $\mathbf{P}$ within a bounded domain: Suppose $\{\mathbf{\hat{x}}_1,\dots, \mathbf{\hat{x}}_N\}$ is a set of quantized vectors in the data space $\mathbb{R}^D$. We say that $\{\mathbf{\hat{x}}_1,\dots, \mathbf{\hat{x}}_N\}$ is a valid quantization strategy of $\mathbf{P}$ regarding the upper bound $U>0$, if for any sample $\mathbf{x}$ drawn from $\mathbf{P}$, there exists at least one quantized vector $\mathbf{\hat{x}}\in\{\mathbf{\hat{x}}_1,\dots, \mathbf{\hat{x}}_N\}$ such that $\|\mathbf{x}-\mathbf{\hat{x}}\|_2\leq U$. Then, the $U$-essence, denoted as $\mathcal{E}_{\mathbf{P},U}$, is the minimum possible size of a valid quantization strategy $\{\mathbf{\hat{x}}_1,\dots, \mathbf{\hat{x}}_N\}$. 

To further analyze the `boundaries' between reconstructed samples from different spiking patterns, we provide the following definition:
\begin{definition}\label{on_boundary}
Suppose $\mathbf{P}$ is a probability distribution generating data sample $\mathbf{x}\in\mathbb{R}^D$ within a bounded domain $\prod_{d=1}^D[\lambda,\mu]$, and suppose $U>0$ is a given upper bound. Suppose $\mathcal{C}_{\mathbf{P},\frac{1}{2}U}^\dagger$ is the set of Spectrum VAEs essentially compatible with $\mathbf{P}$ given $\frac{1}{2}U$. Also, suppose $\mathcal{E}_{\mathbf{P},U}$ is the $U$-essence of $\mathbf{P}$, and suppose $\{\mathbf{\hat{x}}_1,\dots, \mathbf{\hat{x}}_N\}$ is the corresponding set of quantized vectors regarding $\mathcal{E}_{\mathbf{P},U}$.

For a Spectrum VAE $(\phi,\theta)\in \mathcal{C}_{\mathbf{P},\frac{1}{2}U}^\dagger$, we say that two quantized vectors $\mathbf{\hat{x}}_a, \mathbf{\hat{x}}_b\in\{\mathbf{\hat{x}}_1,\dots, \mathbf{\hat{x}}_N\}$ form an \textbf{on-boundary pair} $(\mathbf{\hat{x}}_a, \mathbf{\hat{x}}_b)$ with respect to $(\phi,\theta)$, if $\|\mathbf{\hat{x}}_a - \mathbf{\hat{x}}_b\|_2\leq\frac{1}{2}U$ but $\phi(\mathbf{\hat{x}}_a)$ and $\phi(\mathbf{\hat{x}}_b)$ are preserved by two different spiking patterns of $(\phi,\theta)$. 

We use $\mathcal{B}_{\{\mathbf{\hat{x}}_1,\dots, \mathbf{\hat{x}}_N\}, (\phi,\theta)}$ to denote the number of on-boundary pairs produced by the set $\{\mathbf{\hat{x}}_1,\dots, \mathbf{\hat{x}}_N\}$ with respect to the Spectrum VAE $(\phi,\theta)$.
\end{definition}
Finally, we provide a hypothesis saying that the `boundaries' between reconstructed samples from different spiking patterns are minimized for the essential Spectrum VAE. That is:

\begin{hypothesis}\label{hypo_3}
Suppose $\mathbf{P}$ is a probability distribution generating data sample $\mathbf{x}\in\mathbb{R}^D$ within a bounded domain $\prod_{d=1}^D[\lambda,\mu]$, and suppose $U>0$ is a given upper bound. Suppose $\mathcal{C}_{\mathbf{P},\frac{1}{2}U}^\dagger$ is the set of Spectrum VAEs essentially compatible with $\mathbf{P}$ given $\frac{1}{2}U$, and suppose $\text{MDL}_U^\dagger$ is the corresponding essential minimum description length obtained by formula \ref{format_EMDL} under $\frac{1}{2}U$-complexity. Also, suppose $\mathcal{E}_{\mathbf{P},U}$ is the $U$-essence of $\mathbf{P}$, and suppose $\{\mathbf{\hat{x}}_1,\dots, \mathbf{\hat{x}}_N\}$ is the corresponding set of quantized vectors regarding $\mathcal{E}_{\mathbf{P},U}$. 

Then, if a Spectrum VAE $(\phi^\dagger,\theta^\dagger)\in\mathcal{C}_{\mathbf{P},\frac{1}{2}U}^\dagger$ achieves $\text{MDL}_U^\dagger$, its number of on-boundary pairs $\mathcal{B}_{\{\mathbf{\hat{x}}_1,\dots, \mathbf{\hat{x}}_N\}, (\phi^\dagger,\theta^\dagger)}$ will also be the minimum among all Spectrum VAEs in $\mathcal{C}_{\mathbf{P},\frac{1}{2}U}^\dagger$.
\end{hypothesis}

Once again, we do not provide mathematical proof or further theoretical analysis to Hypothesis \ref{hypo_3}. We only include it in the Appendix, rather than using it to establish our main theory.

If Hypothesis \ref{hypo_3} is correct, then the essential Spectrum VAE $(\phi^\dagger,\theta^\dagger)$ cannot realize reconstruction as described in graph (c). This is because there exists a boundary $x=5$ in the right circle between the samples reconstructed from $\mathcal{P}_2$ and $\mathcal{P}_3$, and such a boundary is irrelative to the data distribution $\mathbf{P}$. Hence, such a boundary adds redundant information to the latent representations, increasing the $U$-redundancy of the Spectrum VAE. As a result, according to Hypothesis \ref{hypo_3}, the essential Spectrum VAE should realize the reconstruction as described in graph (d), where each spiking pattern reconstructs data samples from one entire data class.

\end{document}